\documentclass[lettersize,journal]{IEEEtran}
\usepackage{amsmath,amsfonts}
\usepackage{algorithmic}
\usepackage{algorithm}
\usepackage{array}
\usepackage[caption=false,font=normalsize,labelfont=sf,textfont=sf]{subfig}
\usepackage{textcomp}
\usepackage{stfloats}
\usepackage{url}
\usepackage{verbatim}
\usepackage{graphicx}
\usepackage{cite}
\usepackage{multirow}
\usepackage{multicol}
\usepackage{booktabs}
\usepackage{enumerate}
\usepackage[hidelinks]{hyperref}
\usepackage{xcolor}

\begin{document}

\title{FoodSky: A Food-oriented Large Language Model that Passes the Chef and Dietetic Examination}

\author{Pengfei Zhou,
Weiqing Min,~\IEEEmembership{Senior Member,~IEEE,}
Chaoran Fu,
Ying Jin,
Mingyu Huang,
Xiangyang Li,
Shuhuan Mei,
and~Shuqiang Jiang,~\IEEEmembership{Senior Member,~IEEE}
\thanks{P. Zhou, W. Min, Y. Jin, M. Huang, X. Li and S. Jiang are with the Key Laboratory of Intelligent Information Processing, Institute of Computing Technology, Chinese Academy of Sciences, Beijing 100190, China, and also with the University of Chinese Academy of Sciences, Beijing 100049 China. W. Min and S. Jiang are also with the Institute of Intelligent Computing Technology, Chinese Academy of Sciences, Suzhou 215124, China. C. Fu and S. M are with the Institute of Intelligent Computing Technology, Chinese Academy of Sciences, Suzhou 215124, China, and also with the Zhongke Shenjian (Suzhou) Technology Co., Ltd, Suzhou 215024, China.}
\thanks{E-mail: zpflance@gmail.com, minweiqing@ict.ac.cn, fucr@iict.ac.cn, \{ying.jin, mingyu.huang, xiangyang.li\}@vipl.ict.ac.cn, meish@iict.ac.cn, sqjiang@ict.ac.cn.}}

\markboth{Journal of \LaTeX\ Class Files,~Vol.~X, No.~X, June~2024}%
{Shell \MakeLowercase{\textit{et al.}}: A Sample Article Using IEEEtran.cls for IEEE Journals}


\maketitle

\begin{abstract}

Food is foundational to human life, serving not only as a source of nourishment but also as a cornerstone of cultural identity and social interaction. As the complexity of global dietary needs and preferences grows, food intelligence is needed to enable food perception and reasoning for various tasks, ranging from recipe generation and dietary recommendation to diet-disease correlation discovery and understanding. Towards this goal, for powerful capabilities across various domains and tasks in Large Language Models (LLMs), we introduce Food-oriented LLM FoodSky to comprehend food data through perception and reasoning. Considering the complexity and typicality of Chinese cuisine, we first construct one comprehensive Chinese food corpus FoodEarth from various authoritative sources, which can be leveraged by FoodSky to achieve deep understanding of food-related data. We then propose Topic-based Selective State Space Model (TS3M) and the Hierarchical Topic Retrieval Augmented Generation (HTRAG) mechanism to enhance FoodSky in capturing fine-grained food semantics and generating context-aware food-relevant text, respectively. Our extensive evaluations demonstrate that FoodSky significantly outperforms general-purpose LLMs in both chef and dietetic examinations, with an accuracy of 67.2\% and 66.4\% on the Chinese National Chef Exam and the National Dietetic Exam, respectively. FoodSky not only promises to enhance culinary creativity and promote healthier eating patterns, but also sets a new standard for domain-specific LLMs that address complex real-world issues in the food domain. An online demonstration of FoodSky is available at \href{http://222.92.101.211:8200/\#/home}{\textcolor{red}{http://222.92.101.211:8200}}.

\end{abstract}


\section{Introduction}

Food is fundamental to human survival and culture, closely linked to social values and personal habits~\cite{behrens2017evaluating,asano2019rising}. The rich diversity not only enhances our gastronomic experience but also contributes to the complex world of food data~\cite{mehrabi2021global,basso2020digital}. As society continually evolves, so does the complexity of the food system, leading to the accumulation of vast amounts of data related to culinary practices~\cite{marin2021recipe1m+,damen2020epic}, consumption patterns~\cite{siegrist2020consumer,behrens2017evaluating,popovski2019foodbase} and nutritional content~\cite{thames2021nutrition5k,haussmann2019foodkg}. Food computing has emerged as a key interdisciplinary field towards food intelligence, which leverages this wide range of data to enables critical applications from farm to fork~\cite{min2019survey,grace2021framework,min2023plate}, such as agricultural advice~\cite{ fabregas2019realizing,guo2023wearable, king2017technology}, food robot control~\cite{stella2023can, jones2021bubble, ummadisingu2022cluttered}, culinary creation~\cite{zoran2019cooking,ahnert2013network,ahn2011flavor}, dietary tracking~\cite{althoff2022large,althoff2022large} and chronic disease prevention ~\cite{afshin2019health,zhou2020cmrdf,leng2018network}.

Within the broad spectrum of food computing, cuisine and nutrition stand out as two critical topics due to their direct impact on people's daily lives and well-being. Recent studies have explored various tasks, such as ingredient recognition~\cite{luo2023ingredient, wang2022ingredient, wen2023multi}, recipe retrieval~\cite{marin2021recipe1m+,zhu2019r2gan,salvador2019inverse} and nutrition assessment~\cite{wang2022review,li2023deep,shao2023vision}. For example, ingredient recognition based on the Transformer model has achieved promising results and empowered fine-grained food perception in various applications~\cite{liu2024convolution}. However, the topics of culinary and dietary are complex, as they are closely related to and interact with real-world factors such as cuisine culture and chronic diseases. Previous studies have usually addressed these areas separately, ignoring the potential benefits their integration could bring to food computing research and applications.

Driven by the growth of food systems, the transition from analyzing local data to accumulating massive global datasets is providing new solutions in food computing~\cite{gaupp2021food, min2023large}. This shift supports the implementation of Large Language Models (LLMs), which have demonstrated superior ability to address complex real-world problems in different domains~\cite{hadi2023survey}, such as medicine~\cite{zhou2023survey, yang2024zhongjing}, education~\cite{wang2024large,wardat2023chatgpt}, and finance~\cite{lee2024survey,wu2023bloomberggpt}. Through extensive pretraining and instruction fine-tuning on massive datasets, these models can precisely understand and generate natural language, making them well-suited for tasks involving knowledge and reasoning, such as medical diagnosis~\cite{xu2023baize} and clinical report generation~\cite{zhang2023huatuogpt}. The success in these areas highlights the potential of a food-specific LLM.
The food-specific LLM could leverage massive datasets to deliver precise information perception, understanding and reasoning capabilities tailored to the food domain, addressing critical challenges such as dietary analysis, food recommendations, nutritional diet and cooking advice. This integration of food computing and LLM aims to deliver data-centric insights in food research, enhance user experiences in food applications, and pave one possible way for food intelligence.

Despite the growing interest in LLMs, there have been limited efforts to develop reliable LLMs specifically designed for the culinary and dietetic fields. Recently, some researchers have recognized the potential of LLM in the food domain and have begun applying LLMs in the development of dietary assistants. One notable example is FoodGPT~\cite{qi2023foodgpt}, which proposed a framework to build a knowledge base through a knowledge graph and develop an LLM specific to food. However, FoodGPT has not released a fully-trained model for further development by other researchers. Another study ~\cite{cunningham2023foodgpt} focused on establishing a dietary assistant with the functions of ingredient substitution and recipe recommendation using a food-oriented language model. Furthermore, FoodLMM~\cite{yin2023foodlmm} expands on these efforts by creating a multi-task dietary assistant that recognize food and estimates nutrition simultaneously based on existing multimodal foundation models. These studies highlight the growing interest in using LLMs to address challenges in the food domain and demonstrate the potential for further development of dietary assistants.


Although existing dietary assistants have made significant progress in areas such as multimedia dietary analysis and recommendation, they still have several limitations. First, previous food LLMs were developed based on general language models pre-trained in common scenarios, which can not fully understand and process the fine-grained characteristics of food information, resulting in inaccurate identification and analysis results. Specifically, the lack of certain culinary and dietary knowledge significantly limits their utility in practical food applications, such as ingredient substitution and nutrition assessment. Furthermore, current LLMs have not been adequately tailored to cover the extensive diversity of dietary practices and culinary traditions across various cultures. Recent research has shown that existing LLMs exhibit biases towards Western food knowledge, leading to potentially incorrect or culturally insensitive responses when handling queries from more diverse backgrounds~\cite{zhou2024does}. 

\begin{figure*}[!htbp]
	\centering
	\includegraphics[width=0.99\textwidth]{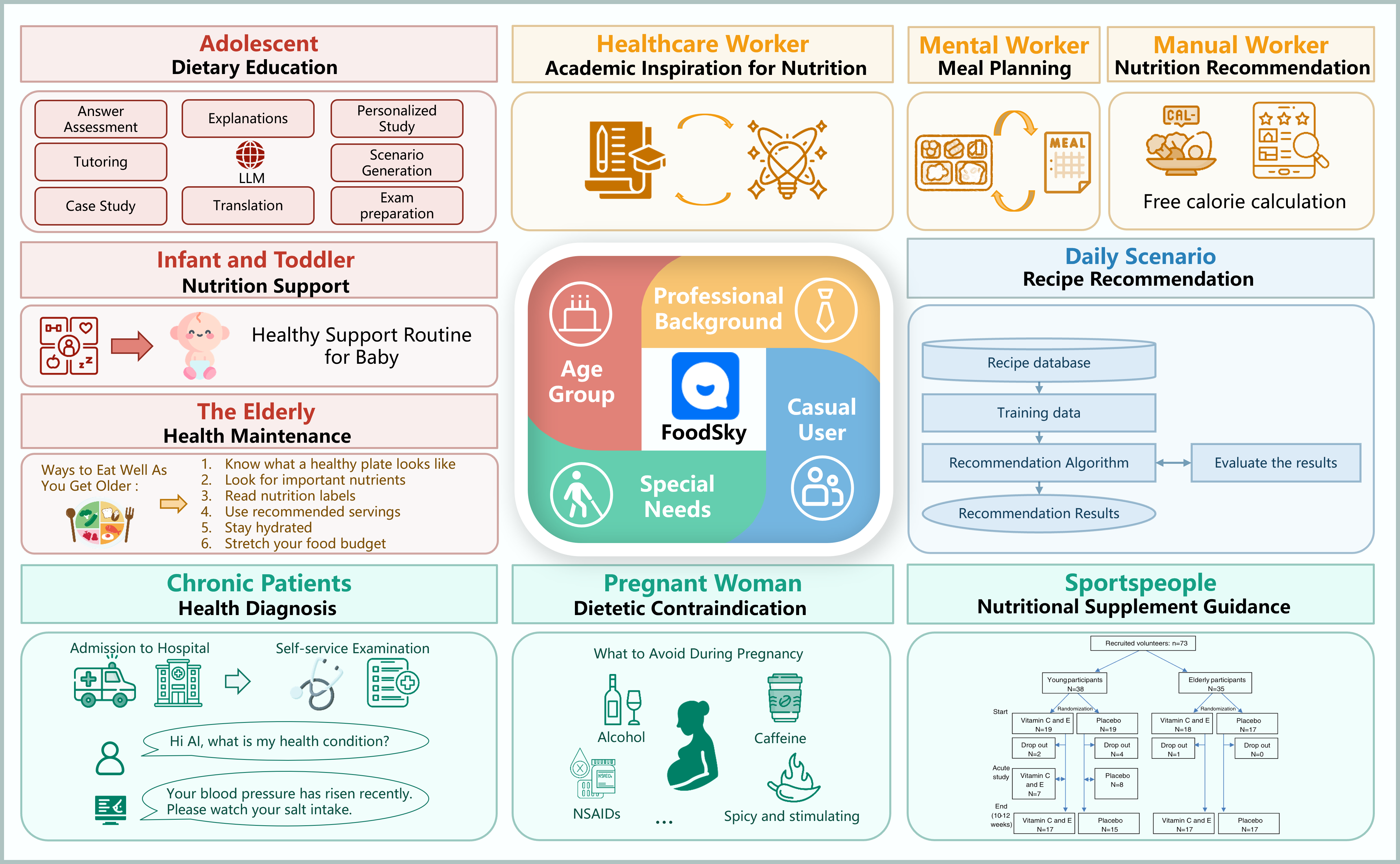}
	\caption{The potential applications of the proposed food-oriented LLM FoodSky in different scenarios.}
	\label{fig:intro}
\end{figure*}

To address these problems, we introduced FoodSky\footnote{As the old Chinese saying goes: food is the most important issue under the sky. Therefore, we name our constructed food-oriented LLM as FoodSky.}, the first Chinese LLM specifically tailored to the food domain. The development of FoodSky, however, faced significant challenges. First, there is a lack of a large-scale food corpus. Unlike fields such as news and media, where data is abundant, food data is relatively limited and scattered across various sources including cooking websites, recipe databases, food blogs, etc. The quality of these data also varies, with spelling and grammar errors, duplicate invalid data, and irrelevant information that complicates the data cleaning process. Second, the food domain covers a wide range of topics including ingredients, cuisines, eating habits, and nutritional information, posing a challenge for the model to understand and handle these diverse topics comprehensively and effectively. Lastly, the cross-cultural nature of food requires the ability to process food-related knowledge across various cultural contexts. Different regions and cultures have distinct eating habits, taste preferences, and cooking traditions, which adds additional complexity for LLMs in handling food queries from various backgrounds.

To overcome these challenges, we first conducted comprehensive data collection and processing to build the first large-scale food-related Chinese corpus, FoodEarth. This corpus contains 811K instruction data from various authoritative sources such as e-books and websites, which were processed through multiple data filtering methods to obtain a high-quality dataset. To enable the model to understand comprehensive and diverse domain knowledge and effectively handle different topic tasks, we developed a Topic-based Selective State Space Model (TS3M) to capture fine-grained food semantics and adapt to diverse topics. Furthermore, we propose a Hierarchical Topic Retrieval Augmented Generation (HTRAG) mechanism to ensure that the model has better generalization and can handle food-related information from different cultural backgrounds based on knowledge enhancement.

Based on the FoodEarth database, TS3M and HTRAG, the proposed FoodSky demonstrates expert-level performance in both chef and dietetic examinations. In particular, FoodSky passes the National Chef Exam and the National Dietetic Examination in China with zero-shot accuracy of 67.2\% and 66.4\%, respectively. More experiments, including both qualitative and quantitative evaluations, further show that FoodSky achieves better performance compared to existing LLMs including InternLM2~\cite{internlm2} and ChatGPT-3.5~\cite{chatgpt} on the Chinese National Chef
Exam and the National Dietetic Exam, demonstrating its ability to provide reliable advice for various culinary and dietetic questions. As shown in Fig.~\ref{fig:intro}, the proposed FoodSky aims to enhance culinary creativity, promote dietary health and benefit diverse groups of people across various scenarios and applications.

\section{Related Work}
This section reviews studies on recent advances related to Large Language Models and food assistant.

\subsection{Large Language Models}

Language is a primary ability in human beings to express and communicate, which develops in early childhood and evolves over a lifetime~\cite{pinker1995language}. However, machines cannot naturally understand and communicate with human language. Therefore, Language Models (LMs) are proposed to enable machines to read and write like humans~\cite{turing2009computing}. LMs model the generative likelihood of word sequences, predicting the probabilities of future tokens. Research on LMs has gone through four stages: Statistical Language Models (SLMs), Neural Language Models (NLMs), Pre-trained Language Models (PLMs), and Large Language Models (LLMs)~\cite{zhao2023survey}. Initially, SLMs focused on enhancing task-specific methods by predicting probabilities and NLMs learned task-agnostic representations to reduce manual feature engineering. Furthermore, PLMs learned context-aware representations through large-scale pretraining. As the latest generation, LLMs stand out based on the scaling effect on model capacity.

Typically, LLMs refer to Transformer models that contain hundreds of billions (or more) of parameters, trained on massive textual data~\cite{shanahan2024talking}, such as GPT-3~\cite{zhou2023chatgpt}, PaLM~\cite{chowdhery2023palm}, Galactica~\cite{taylor2022galactica}, and LLaMA~\cite{touvron2023llama}. The concept of data-centered artificial intelligence (DCAI), initiated by Ng \textit{et al.}~\cite{jakubik2024data}, emphasizes the importance of data over model architecture. This concept is also known as data-centric thinking, where the size and quality of the data are the most crucial factors for developing powerful AI models~\cite{zhao2023survey}. 

Based on the data-centric concept, LLMs are having a significant impact on the AI community, triggering major transformations across multiple research areas~\cite{openai-blog, bubeck2023sparks}. In the field of Natural Language Processing (NLP), LLMs are being used as general-purpose language task solvers. In Information Retrieval (IR), traditional search engines are challenged by AI-driven chatbots. In Computer Vision (CV), researchers are developing GPT-like Multimodal Large Language Models (MLLMs) to address multimodal issues~\cite{huang2024language, cao2023comprehensive, driess2023palm, wu2023visual}. Furthermore, LLMs have been employed in addressing various real-world challenges including medicine~\cite{zhou2023survey, yang2024zhongjing}, education~\cite{wang2024large}, and finance~\cite{lee2024survey}. In medicine, LLMs like ChatGPT have shown potential in medical education~\cite{gilson2023does}, radiologic decision-making~\cite{kung2023performance} and clinical genetics~\cite{duong2024analysis}. Medical-specific LLMs such as Baize~\cite{xu2023baize}, Zhongjing~\cite{yang2024zhongjing} and Huatuo~\cite{zhang2023huatuogpt} are achieving better performance in these medical tasks. In education, LLMs can generate personalized content~\cite{hadi2023survey,gan2023large}, assist with homework~\cite{fraiwan2023review}, and provide real-time feedback in the process of self-study~\cite{wardat2023chatgpt}. In finance, the financial LLMs, including FinMA~\cite{xie2023pixiu}, InvestLM~\cite{yang2023investlm}, FinGPT~\cite{yang2023fingpt}, and BloombergGPT~\cite{wu2023bloomberggpt} are enhancing customer services and providing financial advisory in risk assessment, algorithmic trading and market prediction~\cite{yang2023fingpt}.

Different from these LLMs, FoodSky addresses the unique needs of the food domain through the understanding of the culinary arts and nutritional science. In addition, while general-purpose LLMs may struggle with diverse food topics and cultural backgrounds, the TS3M architecture in FoodSky enables it to capture fine-grained topic differences and the proposed HTRAG ensures the model with more precise and background-aware responses. This capability makes FoodSky stand out as a powerful tool that will influence culinary professionals, dietitians, and consumers, setting a new standard for the application of AI in food-relevant domains.

\subsection{Food Computing}

Food computing is an interdisciplinary field that uses computational methods to address food-related problems in the fields of medicine, biology, gastronomy, and agronomy, thereby playing an important role in academic research and industry applications~\cite{min2019survey}. Academically, food computing brings challenging research topics such as fine-grained recognition to the research community of machine learning~\cite{min2023large,zhou2024synthesizing}. Industrially, food computing enables various key applications such as smart agriculture~\cite{basso2020digital,min2022applications}, automated food processing~\cite{menichetti2023machine,yang2021integrated} and food recommendation~\cite{min2019food,chen2021personalized}.

Among the tasks in the field of food computing, dietary assistant is one important application that can help consumers make smarter and healthier dietary choices~\cite{singh2022conversational,yin2023foodlmm}. It enables intelligent methods in food scenarios like daily cooking, nutrition, and diet health, driven by advances in various research areas such as food recommendation~\cite{min2019food} and recipe retrieval~\cite{marin2021recipe1m+}. Closely related to dietary assistant, food recommendation is important for diet analysis and health management\cite{lin2014content, zhang2016exploiting, cui2023construction, li2023health}. For example, Chu \textit{et al.}\cite{chu2017hybrid} built a hybrid recommendation system by analyzing images in restaurant blogs, and Asani \textit{et al.}\cite{asani2021restaurant} extracted food names from user reviews to analyze dietary feelings. In addition, Ling \textit{et al.}\cite{ling2022following} made recommendations based on behavioral data of users who have successfully lost weight, and Ribeiro \textit{et al.}\cite{ribeiro2018souschef} considered the nutrition, preferences, and budget of the elderly for meal recommendations. Recipe retrieval is another important research topic that provides recipe advice in culinary scenarios\cite{salvador2017learning,salvador2021revamping,wang2021cross,sugiyama2021cross} It is worth noting that Zhu \textit{et al.}\cite{zhu2019r2gan} and Guerrero \textit{et al.}\cite{guerrero2020cross} use GAN to generate images and improve retrieval performance, and Salvador \textit{et al.} \cite{salvador2019inverse} uses a classifier and a Transformer decoder to generate recipe content based on the input images, and obtains higher generalization than conventional retrieval methods.

As LLMs have shown the ability to address complex real-world issues related to knowledge reasoning in various fields, it has been recently explored in the dietary assistant domain. For example, Qarajeh \textit{et al.}\cite{qarajeh2023ai} evaluated the efficacy of different LLMs in identifying potassium and phosphorus levels in foods, and used LLMs to plan healthier diets to prevent complications such as hyperkalemia and hyperphosphatemia. Yin \textit{et al.}~\cite{yin2023foodlmm} proposed a versatile dietary assistant FoodLMM based on the LLaVA~\cite{liu2024visual}. FoodLMM has multi-tasking capabilities for food recognition, ingredient recognition, food segmentation, recipe generation, nutrition estimation and multi-round conversation. Nag \textit{et al.}~\cite{nag2023integrative} proposed a personal health navigation framework that estimates current health status based on personal models, and guides the user towards their healthy goal.

Currently, there is no available comprehensive dataset for food recipe advice, food science popularization, diet recommendations and nutrition assessment. Therefore, we establish the first large-scale Chinese instruction dataset FoodEarth for the food domain. Specifically, unlike previous studies of other domains like medicine~\cite{zhou2023survey, yang2024zhongjing} and education~\cite{wang2024large,lee2024survey}, we adopted a multi-stage dataset construction pipeline that is designed to particularly address the diverse sources and complex raw data problems for Chinese food data. Based on this, the fundamental LLM FoodSky can be built to provide culinary insights and dietary guidance directly for users while boosting the research on dietary assistant and food computing. 
\begin{figure*}[t]
	\centering
	\includegraphics[width=0.98\textwidth]{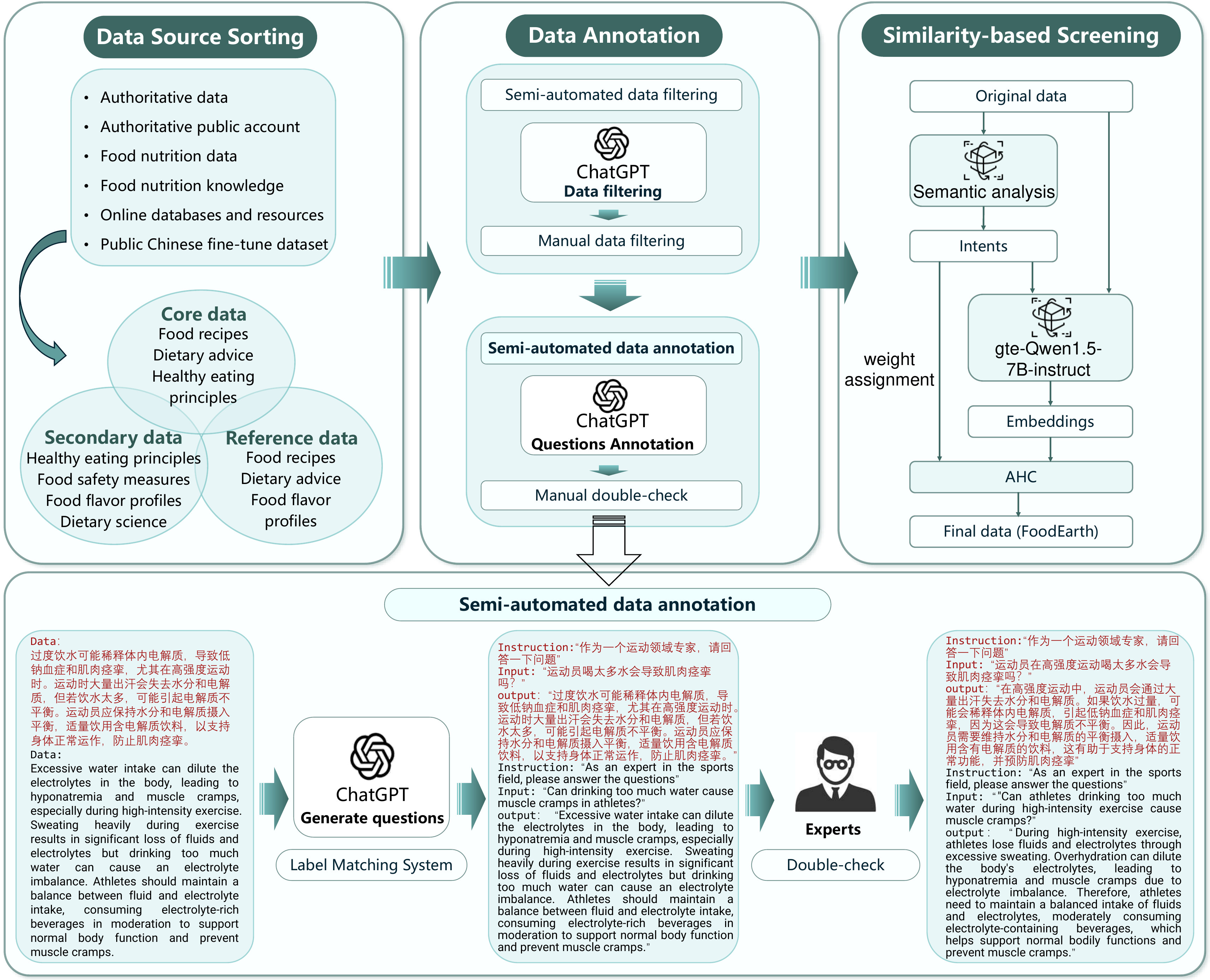}
	\caption{The pipeline of establishing FoodEarth includes data source sorting, data annotation and similarity-based data screening. The semi-automated data filtering and annotation are the main procedures in data annotation. The AHC refers to the Agglomerative Hierarchical Clustering approach.}
	\label{fig:processing}
\end{figure*}

\section{Dataset}

This section details the construction of a large-scale Chinese instruction dataset for the food domain, which is illustrated in Fig.~\ref{fig:processing}. The dataset was compiled from diverse Chinese authoritative sources, including nutrition databases, academic journals, and expert-endorsed websites. To ensure the data quality, we sort the sources of raw data by level, followed by a data annotation process to get the logical instruction data. In the data annotation process, we conduct semi-automated data filtering and semi-automated data annotation. To reduce the repetition rate in the dataset and increase the professionalism of the data, we finally completed the construction of the dataset with experts through similarity-based screening. The established dataset FoodEarth, consisting of 811,491 question-answer pairs, provides a robust foundation for fine-tuning LLMs in the food and nutrition domain.

\subsection{Data Source Sorting}
The main goal of our research is to provide professional consulting services in the domain of cuisine and diet. Our goal was to build a comprehensive knowledge base from six key areas of food expertise: Dietary science, Dietary advice, Food flavour profiles, Food safety measures, Food recipes and Healthy eating principles.

Textual data in the cuisine and diet domain is unique and complex due to its diverse sources. Data in fields like medicine and finance typically exhibit clear cause-and-effect relationships, such as in disease treatment mapping and economic regulation consulting. In contrast, food-related data vary in credibility and often present conflicting opinions. The complexity of cooking, diet, nutrition and science data arises from individual taste preferences, variations in food measurements, regional food diversity and different cooking methods.


To address these challenges, our textual dataset is built based on a wide range of authoritative sources, enhancing its reliability and comprehensiveness and providing a detailed understanding of this field. This diversity is crucial, as it captures the range of cooking techniques and personalized nutrition, making it different from datasets in other fields. The combination of information from various origins ensures that our dataset models the unique characteristics of the cuisine and diet domain, focusing on specific food challenges.

\begin{figure*}[t]
	\centering
	\includegraphics[width=1.01\textwidth]{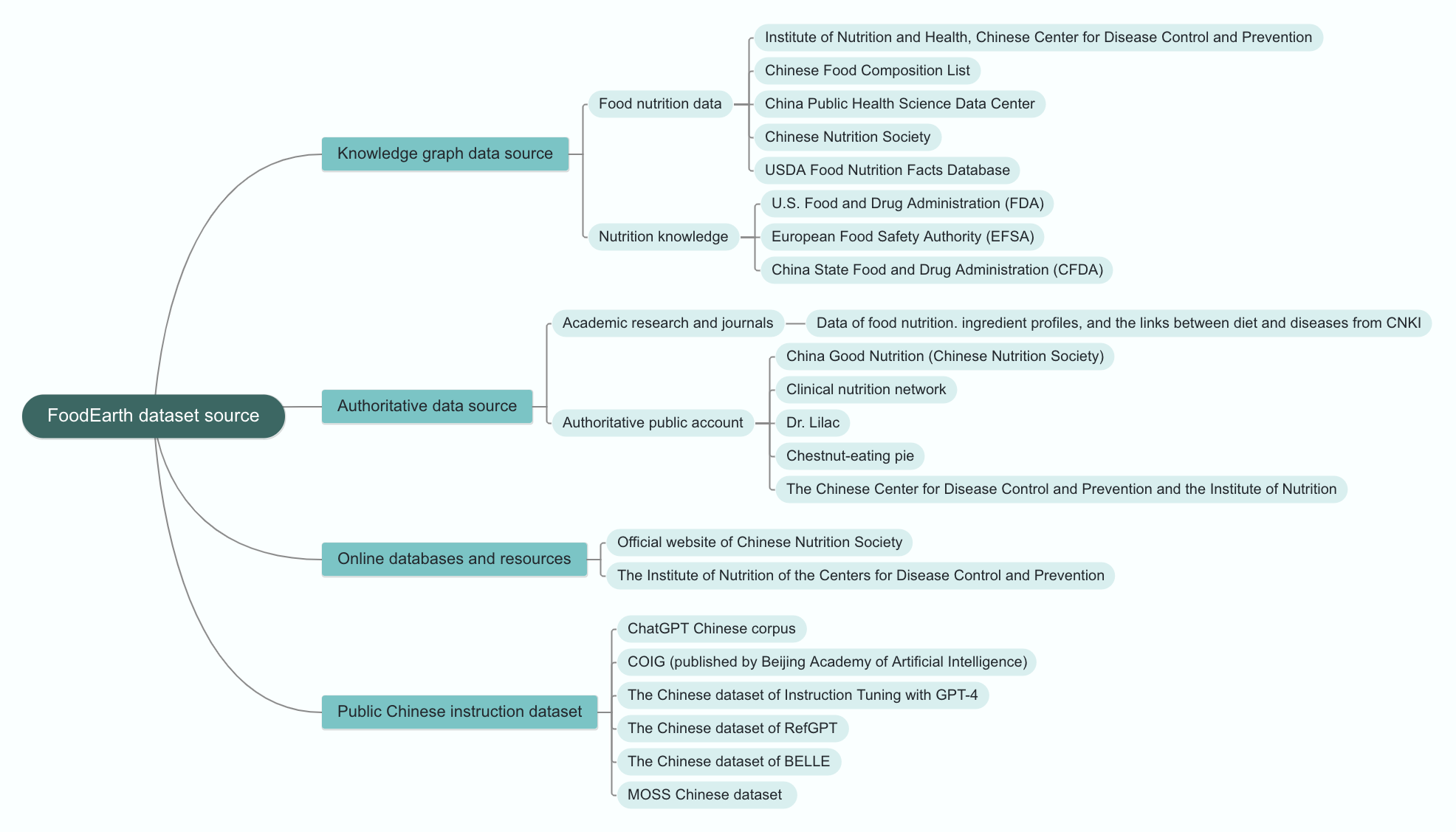}
	\caption{The illustration of authoritative data sources of our FoodEarth.}
	\label{fig:source}
\end{figure*}

Currently, mainstream food data widely exists in professional books, authoritative nutrition papers, authoritative websites, public accounts and other online channels. As shown in Fig.~\ref{fig:source}, we divided the data sources into knowledge graph data source, authoritative data source, online databases and resources, and public Chinese instruction dataset, and then reorganized the data according to the authority and timeliness of the data.

In the food field, we accumulated millions of food nutrition and recipe knowledge graph data nodes through the above-mentioned authoritative website content as the core data content of this study. CNKI professional papers and authoritative websites in the field serve as secondary data sources including dietary recommendations, healthy eating principles, food safety measures, food flavor profiles and dietary science professional data. In addition, we also compiled other resources and generated data related to food and beverages through ChatGPT and public Chinese datasets as reference datasets to enrich our data scope. In the end, through multiple rounds of manual annotation, professional filtering, ChatGPT data processing and other data cleaning methods, we collected a pure text dataset of more than 2.5 billion tokens and completed the construction of a Q\&A dataset with 811,491 pairs.

\begin{table*}[htbp]
\caption{Data volume proportion and data content of each data source}
\renewcommand{\arraystretch}{1.5}
\label{tab:model-comparison}
\centering
\begin{tabular}{lccccc}
\hline
Source Origin & Raw data & Instruction & Modalities & Data Types & License \\
\hline
Authoritative public account & 25,482,971 tokens & 8,401 pairs & text &news,article & CC BY-SA 3.0 \\
Authoritative Data & 22,680,536 tokens & 12,424 pairs & text & report, academic papers, book & CC BY-SA 3.0 \\
Knowledge graph data & 6,480,672 tokens & 538,376 pairs & structured data & knowledge graph & CC BY-NC 3.0 \\
Public Chinese fine-tune dataset tokens & 448,774,454 tokens & 253,986 pairs & structured data & fine-tune dataset & Publish \\
\hline
\end{tabular}
\end{table*}

\subsection{Data Annotation}

In the process of constructing a high-quality dataset, we employed a semi-automated strategic combining automated processing and manual processing to ensure the quality and practicability of our established FoodEarth. Specifically, our processing procedure can be divided into two stages: semi-automated data filtering and semi-automated data annotation. During the semi-automated data filtering stage, we utilized the semantic understanding capabilities of ChatGPT to analyze the raw Chinese public question-answer dataset and identify food-related content, which reduced the workload of manual filtering. In the data annotation stage, we employed the ChatGPT API to generate diverse questions based on the filtered food-related raw data. Our annotation team then thoroughly reviewed each generated question and answer (Q\&A) pair, validating the accuracy of the questions and answers. Any inaccurate or ambiguous Q\&A pairings are improved or replaced to meet our strict standards. At each stage, we utilize ChatGPT to enhance efficiency, while incorporating manual verification and refinement by our expert annotation team to ensure accuracy.

\subsubsection{Semi-automated Data Filtering}
During the semi-automated data filtering process, we completed the work in three stages due to the wide variety of food text datasets, the wide range of sources, and the uneven levels of data quality. Initially, our team performed manual filtering based on pre-set quality criteria and relevance metrics. For example, we first filtered out noise data such as irrelevant modal words and emoticons. This stage involved ten team members with data processing experience using the data mining system simultaneously. It took nearly a month to complete the preliminary data cleaning and store it in a dictionary format as the initial filtered data to facilitate subsequent processing.

\begin{figure*}[t]
	\centering
	\includegraphics[width=0.9\textwidth]{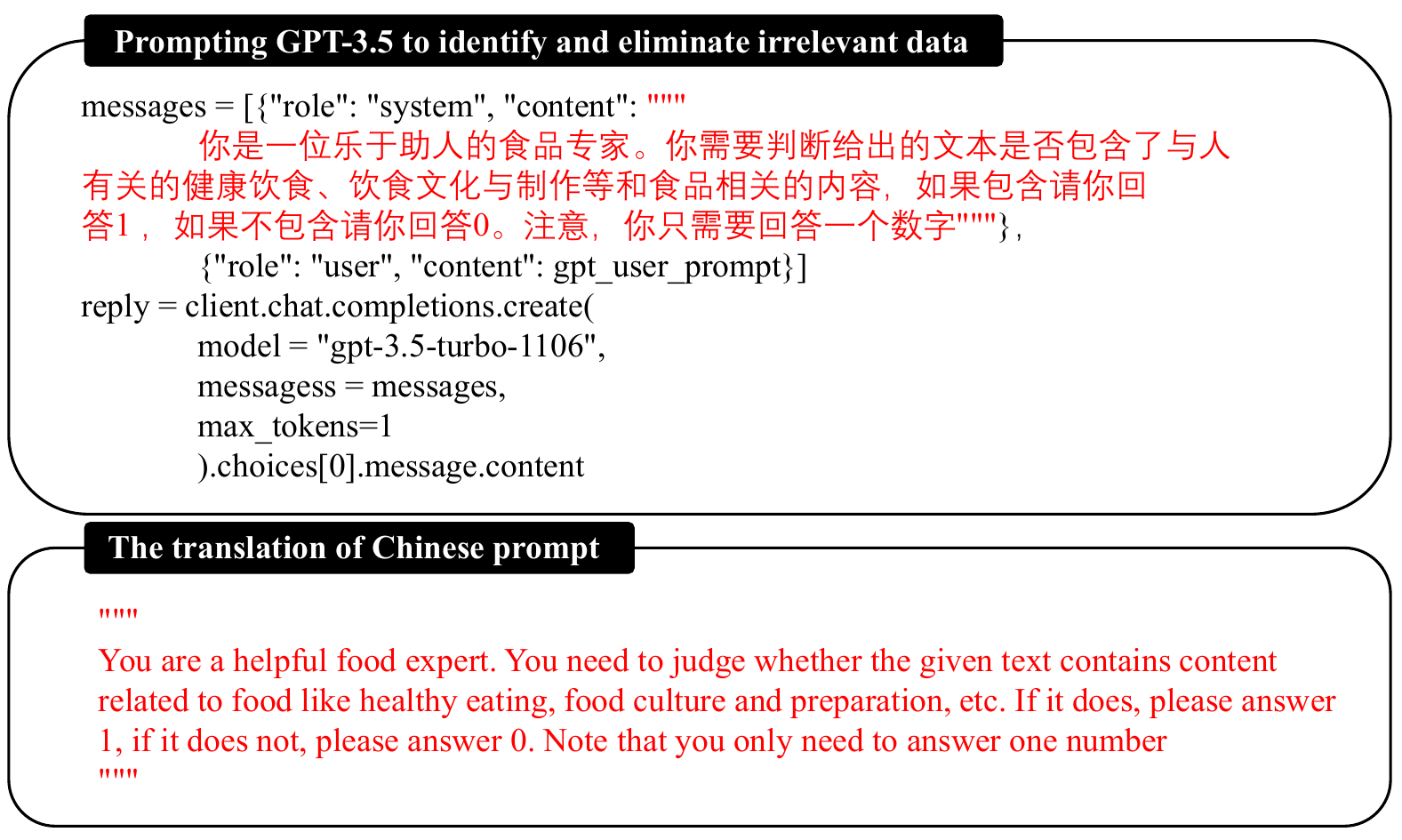}
	\caption{The prompt used to automatically filter raw data based on ChatGPT in our semi-automated data filtering procedures.}
	\label{fig:prompt}
\end{figure*}

Based on the manual filtering, we further enhanced the data filtering process by leveraging the ChatGPT API. As shown in Fig.~\ref{fig:prompt}, we set up a prompt mechanism to identify and remove irrelevant data, requiring ChatGPT to return 0 for text data that did not belong to the research topic as a negative sample, and return 1 for relevant text data as a positive sample. We then used the filtering system to judge the return value, filtering out the negative sample data and retaining only the positive sample data. We recorded this as secondary filtered data, which can reveal fine-grained correlation patterns that human filters may have missed.

After processing with ChatGPT, we submitted the dataset to three domain experts with deep expertise in food safety and nutrition. These experts conducted a comprehensive double-check on the filtered dataset, verifying its relevance and assessing the authenticity and trustworthiness of the data. They ensured the accuracy and usefulness of the dataset by comparing it with the latest literature and industry reports, ensuring fully cleaned final data. This step is important throughout the data preparation process, as the quality of the data may ultimately influence the health and physical well-being of users who rely on the information provided by the fine-tuned LLM.

\subsubsection{Semi-automated Data Annotation}
To effectively refine an LLM, we make more efforts to construct a comprehensive text-based question-answering (Q\&A) training dataset, specifically tailored to the food domain. We carried out a dual process of automated generation and manual verification to ensure the quality of the dataset.

We used the ChatGPT API to automatically generate food-related question-answer pairs. By employing prompt engineering techniques, we carefully designed prompts that guided the model to create a diverse set of questions closely reflecting the information in the text. These prompts were iteratively refined to ensure that the generated questions covered a broad range of information needs, from basic definitions to complex conceptual discussions. The prompts played a crucial role in directing the model to produce high-quality, relevant questions that aligned with the intended scope and depth of the dataset. After completing the construction of the question through ChatGPT, we used a simplified matching algorithm to accurately extract the text paragraph corresponding to the question from the dataset after completing the data filtering.

With the automated generation phase complete, our experienced annotation team conducts a thorough secondary inspection. This team, consisting of food science experts, engaged in a detailed review of each Q\&A pair, validating the pertinence of the questions and the accuracy of the answers. Throughout the annotation process, any Q\&A pairs deemed imprecise or ambiguous were directly refined or replaced to meet our exacting standards.

\begin{figure*}[t]
	\centering
	\includegraphics[width=1.01\textwidth]{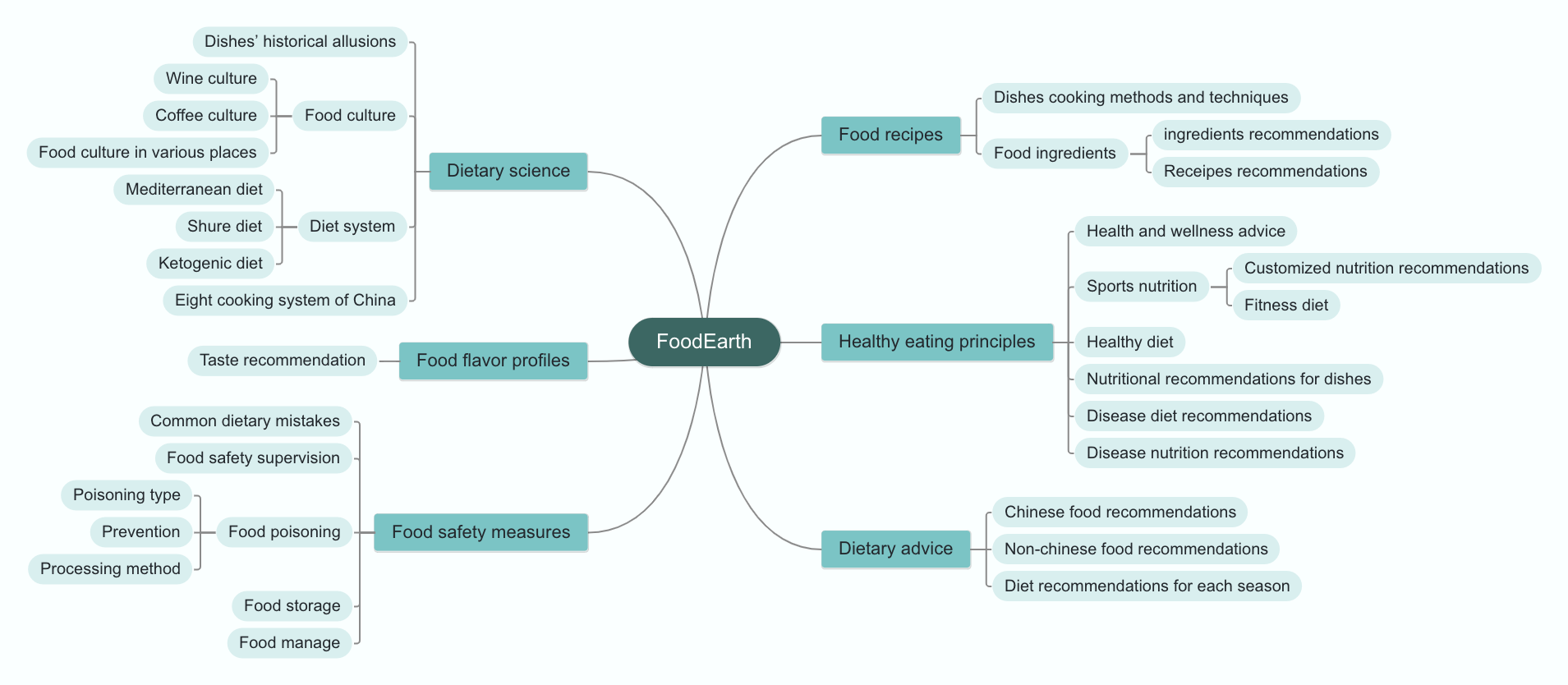}
	\caption{Hierarchical structure of topics in our FoodEarth dataset.}
	\label{fig:topic}
\end{figure*}

To further improve the quality of the established dataset, we carefully documented and reviewed the workflow of the annotation team to identify and reduce potential biases or errors. This thorough review process was essential in establishing a high-quality, text-based question-answering training dataset specifically for the food domain. As shown in Fig.~\ref{fig:topic}, the established dataset FoodEarth covers a wide range of food-related topics and has undergone strict quality assurance measures. As a result, FoodEarth can provide a strong foundation for fine-tuning the Chinese language model, ensuring that it is ready for complex, domain-specific applications.

\subsection{Similarity-based Screening}
\label{subsec:Similarity-Based Screening}

After constructing FoodEarth, a large-scale food dataset that contains almost 1 million pairs of instruction data, we found that there are many semantically similar question and answer pairs, which made our model train on data that was repeated many times and failed to achieve good generalization performance. Therefore, to effectively process and deduplicate this large-scale dataset, we consider using the existing text representation model to encode the text for question and answer, and then use similarity to perform deduplication operations. We first sampled a mini-test set containing 506 pieces of data to evaluate and select the encoding model that best suits our similarity calculation task. We experimented with multiple text representation models on this mini-test set and selected the best-performing model for encoding the full dataset.
Then we apply the selected encoding model to the entire dataset, transforming each piece of data into an encoding vector. By comparing the similarity of feature vectors between different data items, we use threshold setting and clustering methods to judge and identify duplicate data, and delete any data items with a similarity score higher than 0.9 with other centers of clustering in the dataset. Ultimately, this approach effectively screens out the high-quality data by reducing data redundancy, which ensures the generalization performance of the proposed model.

\subsubsection{Mini-Test Set}
\label{subsubsec:mini-test-set}
To determine the method of similarity-based screening that best fits our dataset, we first built a mini-test set to measure the performance of different embedding methods. The mini-test set consists of 506 samples, with two different sentences and a 0/1 label, representing whether the two texts are similar.

\subsubsection{Embedding Methods}
\label{subsubsec:embedding-methods}
To determine the text similarity calculation method suitable for our dataset, and to better realize the similarity filtering task, we selected several models with the best text embedding performance at present and tested them on the mini-test set. We adopted several methods that currently rank high on C-MTEB, including gte-base-zh\cite{li2023towards}, gte-large-zh\cite{li2023towards}, gte-Qwen1.5-7B-instruct\cite {li2023towards}, bge-large-zh-v1.5\cite{bge_embedding}, acge\_text\_embedding\cite{intsigacge}, Baichuan-text-embedding\cite{yang2023baichuan}, stella-mrl-large-zh-v3.5\cite{infgradstella} and puff-base-v1\cite{infgradpuff}: 
\begin{itemize}
\item gte-base-zh \cite{li2023towards}: One of the GTE series models in Chinese, using the Dual Encoder framework. The model was initialized using the base model of the BERT model with 110M parameters.
\item gte-large-zh \cite{li2023towards}: One of the GTE series models in Chinese, using the Dual Encoder framework. The model was initialized using the large model of the
BERT model with 330M params.
\item gte-Qwen1.5-7B-instruct \cite{li2023towards}: This model has been engineered starting from the Qwen1.5-7B LLM, drawing on the robust natural language processing capabilities of the Qwen1.5-7B model.
\item bge-large-zh-v1.5 \cite{bge_embedding}: BAAI General Embedding is a series of large embedding models open-sourced by Beijing Zhiyuan Artificial Intelligence Research Institute, referred to as BGE, and supports Chinese and English embedding. BGE-Large-zh is the Chinese vector large model with the largest parameter scale in the BGE series, with 326 million parameters. Input sequence 512, output dimension 1024.
\item acge\_text\_embedding \cite{intsigacge}: Compared with traditional pre-training or fine-tuning vertical domain models, the acge model supports the construction of general classification models in different scenarios, improves the accuracy of long document information extraction and has a relatively low application cost. It mainly uses the representation learning framework of Matryoshka Representation Learning (MRL).
\item Baichuan-text-embedding \cite{yang2023baichuan}: The embedding model is self-developed by Baichuan Inc. It is pre-trained on high-quality Chinese data of more than 1.5T tokens and uses a self-developed loss function to solve the problem of contrastive learning methods relying on batch size.
\item stella-mrl-large-zh-v3.5 \cite{infgradstella}: This model is trained using the MRL method based on stella-large-zh-v3-1792d. Its main feature is variable vector dimensions.
\item puff-base-v1 \cite{infgradpuff}: The Puff series of models are specifically designed for retrieval and semantic matching tasks, and pay more attention to generalization and the effect of private general test sets. The vector dimension of this model is variable and supports bilingualism in Chinese and English.
\end{itemize}


\subsubsection{Comparison of Different Embedding Methods}
\label{subsubsec:evaluation-metrics}
To compare the performance of each model in terms of similarity calculation, we tested the above models on the mini-test set we constructed. The evaluation metrics used to assess the performance of the embedding methods include accuracy and F1. These metrics are chosen because they effectively capture the similarity estimation performance and are widely used in the related literature.

\begin{equation}
\begin{aligned}
 Accuracy=\frac{\left|TP \right|+\left|TN\right|}{\left|TP\right|+\left|TN\right|+\left|FP\right|+\left|FN\right|}
 \end{aligned}
\end{equation}

\begin{equation}
\begin{aligned}
F1-score =2\times \frac{Precision\times Recall}{Presion+Recall}
\end{aligned}
\end{equation}


We conducted experiments on several models on the mini-test set, and the experimental results are shown in Table~\ref{tab:model-comparison}. The results show that each model shows excellent performance on the similarity assessment task. The accuracy rates all exceeded 98\%, and only a few data were judged incorrectly, showing that existing text embedding models can understand the semantics of the food field well. Besides, The results show that the gte-Qwen1.5-7B-instruct model outperforms the other embedding methods on the mini-test set, achieving a 99.41\% score of accuracy and a 99.41\% score of F1. This indicates that the model is more suitable for capturing the semantic similarities between food items and distinguishing different semantics about food. Therefore, we selected this model for similar data filtering.

\begin{table}[htbp]
\caption{Comparison of Similarity Estimation Performance of Different Embedding Methods (\%)}
\setlength{\tabcolsep}{10pt}
\renewcommand{\arraystretch}{1.5}
\label{tab:model-comparison}
\centering
\begin{tabular}{lcc}
\hline
Methods & Accuracy & F1 \\
\hline
gte-base-zh & 99.01 & 99.02 \\
gte-large-zh & 98.81 & 98.81 \\
bge-large-zh-v1.5 & 99.01 & 99.00 \\
acge\_text\_embedding & 98.02 & 98.02 \\
Baichuan-text-embedding & 99.21 & 99.20 \\
stella-mrl-large-zh-v3.5 & 99.01 & 99.02 \\
puff-base-v1 & 97.43 & 97.50 \\
\textbf{gte-Qwen1.5-7B-instruct} & \textbf{99.41} & \textbf{99.41} \\
\hline
\end{tabular}
\end{table}
\subsubsection{Data Screening}
\label{subsubsec:data-filtering-process}
Based on the evaluation results, we use the gte-Qwen1.5-7B-instruct embedding method for data filtering and further clustering processing. 
We first tried to filter similar data by setting a threshold. This method first uses the gte-Qwen1.5-7B-instruct embedding model to obtain the embedding representation of all question and answer pairs in the dataset. Then we calculate the cosine similarity between each two data. We follow the threshold setting of experiments on the mini-test set and set the threshold to 0.9. When the value of cosine similarity is greater than the threshold, we regard the two pieces of data as similar data and remove the shorter data. However, this method is often inaccurate for many data whose text content is similar but the actual intention is different. Therefore, we adopted a clustering method that jointly embedded representations of intentions and performed similarity filtering on the dataset. The process is given as follows: 

\begin{enumerate}
\item Advanced language models are used to perform semantic analysis on question texts, identify their core intents and classify them into corresponding intent categories. These intent categories not only describe the topic of the question but also reflect its underlying user needs and goals.

\item The extracted intents to quantify the frequency and weight of various intents in the dataset are modeled to guide the subsequent clustering process. The model ensures that we fully understand the diversity and prevalence of intents through statistical analysis and probability distribution.

\item The processed question-answering data and its corresponding intent categories are input into the gte-Qwen1.5-7B-instruct text embedding model, which uses advanced text embedding technology to convert text into a high-dimensional vector representation, retain the original semantic information and reflect the semantic similarity between texts.

\item Agglomerative Hierarchical Clustering (AHC) is used to classify the data, after obtaining the embedding representation. AHC is a bottom-up method that forms a hierarchical structure by iteratively merging the most similar data points. We incorporate the intent distribution information into the clustering algorithm and ensure that the clustering results are based on both semantic similarity and the importance of each intent category through weight assignment. The weights are adjusted according to the probability and frequency in the intent distribution model, making data points of common intent categories more influential in the clustering.

\item The formation of the final data set depends on the selection of representative data points. This process takes into account the centrality of the data points in the micro-category to ensure that the data set is diverse and representative. The distance between each data point and the centre of the micro-category to which it belongs is calculated, and the data point with the smallest distance is selected to fully demonstrate the characteristics of each category.

\end{enumerate}

After completing the above steps, the dataset has been optimized to improve the quality and diversity of the dataset. We invite two experts to conduct secondary validation on this refined dataset to ensure the usability of each piece of final, thus resulting in the final version of FoodEarth with 811,491 Chinese instruction data, providing a solid foundation for further research and application.


\begin{figure*}[!htbp]
	\centering
	\includegraphics[width=0.94\textwidth]{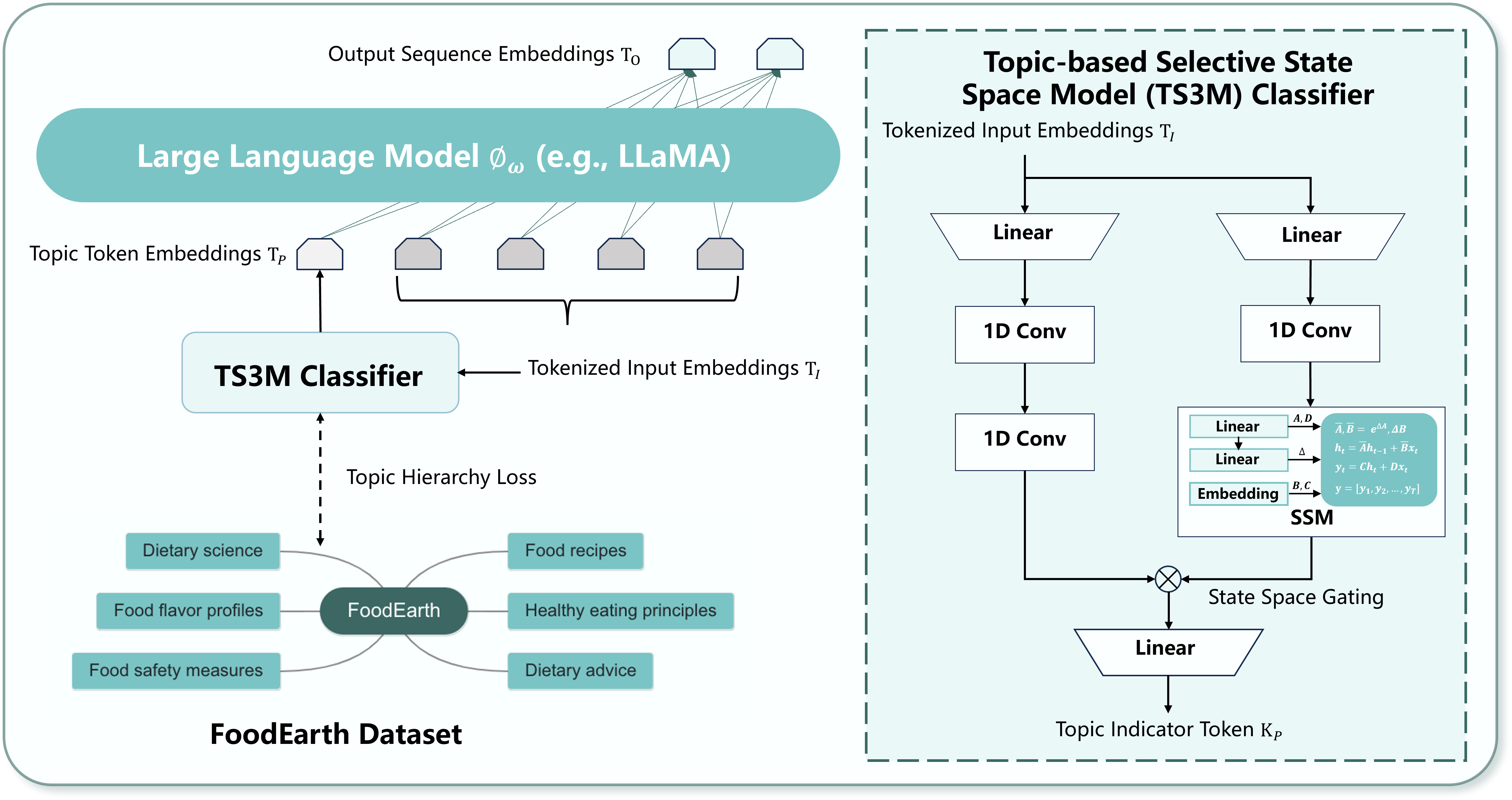}
	\caption{The core model of the proposed FoodSky method. During the inference, the Hierarchical Topic Retrieval Augmented Generation (HTRAG) module uses the topic indicator token for food knowledge retrieval.}
	\label{fig:method}
\end{figure*}

\section{Method}

In FoodSky, we propose two key algorithms, Topic-based Selective State Space Model (TS3M) to enhance the accuracy of the model by integrating topic-relevant information and Hierarchical Topic Retrieval Augmented Generation (HTRAG) to enable the generation of responses that are richer in informational content. The backbone model and instruction-tuning are first introduced to provide a technical background for the proposed algorithm.




\subsection{Backbone Model}
The backbone model serves as the foundation for our proposed extensions, including the Topic-based Selective State Space Model (TS3M) and Hierarchical Topic Retrieval Augmented Generation (HTRAG), which can further enhance the understanding ability of the proposed model and generate responses to food-related instructions and questions.

Specifically, our backbone model $\phi_\omega$ is an LLM (e.g. LLaMA-2) pretrained on the large-scale Chinese corpus and finetuned on Chinese instruction dataset from the general domain. We name this backbone as Chinese LLaMA-2 (CLLaMA2). To adapt the model to the specific challenges and characteristics of the food domain, we conduct food-specific finetuning. This stage involves training the model on the proposed FoodEarth dataset, which consists of food-related Chinese instructions, questions, and answers. The data covers various aspects of the food domain, including recipes, ingredients, nutrition, cuisines, and food safety. By focusing on food-specific data, the model learns to generate more accurate and informative responses to food-related queries.

\subsection{Instruction Fine-tuning}

Instruction fine-tuning plays a crucial role in adapting our backbone language model to follow instructions effectively and generate appropriate responses in the food domain. We employ an instruction fine-tuning approach with two stages: general instruction tuning and food-specific instruction tuning. Through the first stage of general instruction tuning on the large-scale general-domain dataset, we obtained the instructed-tuned backbone CLLaMA2-Alpaca. This model has the ability to understand and follow instructions in context, providing basic instruction-following capabilities.

After the general instruction tuning stage, the model has obtained a solid foundation in instruction-following. However, to specialize in the food domain, we proceed to the second stage: food-specific instruction tuning. In this stage, we further fine-tune the model using the food-related instructions and their corresponding responses in our proposed FoodEarth. This dataset covers a wide range of food-related topics, such as dietary science, cooking techniques, healthy eating principles, and food safety guidelines. By focusing on food-specific instructions, the model learns to understand and generate responses that are tailored to the unique terminology and challenges of the food domain, taking into account factors like ingredient properties, cooking methods, and dietary restrictions. To ensure the effectiveness of the food-specific instruction tuning, we employ several strategies:

\begin{itemize}
\item  Data quality control: We carefully collect, process and double-check the FoodEarth dataset to ensure its accuracy, relevance, and diversity. This involves collecting data from reliable sources, such as culinary experts, food science publications, and reputable cooking websites. We also perform multiple data cleaning and preprocessing procedures involving an annotation team, automated methods, and experts to remove irrelevant or low-quality examples.
\item  Training techniques: During the general instruction tuning stage, we employ techniques such as masked language modeling (MLM) and next sentence prediction (NSP) to enhance the model's understanding of instructions and their relationships to the corresponding responses. MLM involves randomly masking a portion of the input tokens and training the model to predict the masked tokens based on the surrounding context. This helps the model learn to capture the semantic and syntactic patterns in the instructions. NSP, on the other hand, trains the model to predict whether a given response follows logically from the preceding instruction, improving its ability to generate coherent and relevant responses.
\item  Iterative fine-tuning: We employ an iterative fine-tuning approach, where we gradually increase the complexity and specificity of the food-related instructions during training. This allows the model to progressively adapt to the fine-grained characteristics and challenges of the food domain, improving its performance with more specialized capability.
\end{itemize}

\subsection{TS3M Classifier}

Fig.~\ref{fig:method} also presents the Topic-based Selective State Space Model (TS3M). TS3M is designed to capture the semantic relationships between topic and content in the input instructions, enhances the output with prior knowledge of specific food topics, and enables the model to generate more coherent and informative responses.

TS3M takes the tokenized instruction embeddings $\mathbf{T}_I \in \mathbb{R}^{n\times d_m}$ as input, where $n$ represents the number of tokens and $d_m$ is the embedding dimension of tokens. It employs a two-branch architecture consisting of a content representation branch and a topic state representation branch. The content representation branch combines linear and convolutional layers to refine semantic embeddings progressively through multiple layers. The topic state representation branch is designed to influence the output of the model with the selectively integrated internal state. 

On the content representation branch, the initial layer first transforms the input token embeddings to adapt their dimensionality for convolutional processing:
\begin{equation}
\mathbf{X}_L^1 = \mathbf{T}_I \mathbf{W}^1 + \mathbf{b}^1,
\label{eq:linear}
\end{equation}
where $\mathbf{W}^1 \in \mathbb{R}^{d_m\times d_l}$ and $\mathbf{b}^1 \in \mathbb{R}^{d_l}$ represent the weights and biases of the linear transformation, the dimension $d_l$ of the projected input. After the linear transfomation, we obtain the projected input $\mathbf{X}_L^1 \in \mathbb{R}^{n \times d_l}$. Furthermore, a series of one-dimensional convolutional (1D Conv) layers refine these embeddings based on context relationships among tokens.




Specifically, two 1D Conv layers are implemented to model the relationships among tokens and downsample features:
\begin{equation}
\mathbf{X}_{C2}^1 = \sigma ( \text{Conv} ( \mathbf{X}_{C1}^1) ) = \sigma (\text{Conv} ( \sigma ( \text{Conv} ( \mathbf{X}_L^1) ) ) ),
\end{equation}
where $\mathbf{X}_{C1}^1 \in \mathbb{R}^{n \times d_c}$ and $\quad \mathbf{X}_{C2}^1 \in \mathbb{R}^{n \times d_c}$ are the output of 1D Conv layers, $\text{Conv}(\cdot)$ denotes the 1D convolutional operation and $\sigma(\cdot)$ denotes the non-linear operation of LeakyReLU~\cite{maas2013rectifier}.

On the topic state representation branch, TS3M incorporates the State Space Model (SSM) as its key technique, as outlined in Fig.~\ref{fig:method}, to selectively update the latent state based on the instruction input sequences projected through a linear transformation and 1D convolutional operation. The adjustment of the latent state is indirectly guided by the projected instruction token during the training, which can be described as:
\begin{equation}
\mathbf{h}(t) = \mathbf{\hat{A}} \mathbf{h}(t-1) + \mathbf{\hat{B}} \mathbf{x}(t),
\label{eq:ssm1}
\end{equation}
where $\mathbf{h}(t)$ is the latent state at time $t$, $\mathbf{x}(t)$ denotes the projected instruction tokens at time $t$. $\mathbf{A}$ and $\mathbf{B}$ are matrices that define the state dynamics and input influence respectively. $\mathbf{\Delta}_t$ is a timescale parameter.

The state transition is governed by the following procedures, which utilize an exponential matrix to update the state dynamically:
\begin{equation}
\mathbf{\hat{A}} = e^{\mathbf{\Delta}_t\mathbf{A}},
\end{equation}
\begin{equation}
\mathbf{\hat{B}} = (\mathbf{\Delta}_t\mathbf{A})^{-1}(e^{\Delta_t\mathbf{A}} - \mathbf{I})\cdot\mathbf{\Delta}_t\mathbf{B},
\end{equation}

Specifically, TS3M leverages Eq.~\eqref{eq:ssm1} to update the latent state $\mathbf{h}(t)$ by combining the previous state $\mathbf{h}(t-1)$ with the current input $\mathbf{x}(t)$ through a linear transformation. This allows TS3M to selectively incorporate relevant information from the input instructions into the latent state based on the learned matrices $\mathbf{A}$ and $\mathbf{B}$. This approach ensures that the model captures the most relevant information aligned with the topic of the input and generates accurate responses with improved quality and relevance for food and dietary advice tasks.

The latent state $\mathbf{h}(t)$ is further mapped to an output representation $\mathbf{y}(t)$ through another linear transformation defined by matrices $\mathbf{C}$ and $\mathbf{D}$. This output representation captures the semantic information relevant to the current topic and can be used for the downstream generation task. The output generation is described by:
\begin{equation}
\mathbf{y}(t) = \mathbf{C} \mathbf{h}(t) + \mathbf{D} \mathbf{x}(t),
\label{eq:ssm2}
\end{equation}
where $\mathbf{y}(t)$ represents the output at time $t$. $\mathbf{C}$ and $\mathbf{D}$ are matrices that map the latent state and input to an output space respectively.

The output sequence of topic state representation is constructed from the individual outputs over the sequence:
\begin{equation}
\mathbf{y} = [\mathbf{y}_1, \mathbf{y}_2, \dots, \mathbf{y}_T],
\end{equation}
where $\mathbf{y}_t$ corresponds to the output at each time step $t$, and $T$ is the total number of time steps.

The integration of the content representation with the topic state representation is performed through state space gating $\text{Gate}(\cdot)$ to obtain the output topic sequence $\mathbf{T}_s \in \mathbb{R}^{n \times d_s}$, where $d_s$ is the projected dimension.

\begin{equation}
\text{Gate}(\cdot) = \sigma(\mathbf{W}_{G1} \mathbf{X} + \mathbf{W}_{G2} \mathbf{Y}),
\end{equation}
where $\mathbf{W}_{G1} \in \mathbb{R}^{d_s \times d_s}$ and $\mathbf{W}_{G2} \in \mathbb{R}^{d_s \times d_s}$. $\mathbf{X} \in \mathbb{R}^{n \times d_s}$ and $\mathbf{Y} \in \mathbb{R}^{n \times d_s}$ represent the content representation and topic state representation, respectively. This process ensures that the output is context-relevant and topic-focused, which provides the response about the specific food topic addressed in the input instruction. Finally, an indicator token encoder with a linear layer and a pooling layer downsamples the $\mathbf{T}_s$ into the output topic indicator token:

\begin{equation}
    \mathbf{K} = \text{GMP}(\mathbf{T}_s\mathbf{W}_o + \mathbf{b}_o),
\end{equation}
where $\mathbf{K} \in \mathbb{R}^{1 \times P}$ and $P$ is the topic numbers in the dataset. $\text{GMP}$ represents a global maximum pooling operation, $\mathbf{W}_o \in \mathbb{R}^{d_p \times d_o}$ represents the transformation matrix, $\mathbf{b}_o \in \mathbb{R}^{d_o}$ represents the bias.

To combine with the reasoning procedure of the LLM, the upsampled topic token embeddings $\mathbf{T}_s \in \mathbb{R}^{1 \times d_m}$ is concatenated with the tokenized input embeddings $\mathbf{T}_I \in \mathbb{R}^{n \times d_m}$. The combined embeddings work as the final input for the LLM during the training and inference. The topic token embeddings upsampled from the topic indicator token $\mathbf{K}$ provide the topic prior for generating responses.

TS3M offers enhanced capability in generating context-aware textual responses using topic knowledge from the FoodEarth Dataset. Moreover, TS3M can be integrated with the Hierarchical Topic Retrieval Augmented Generation (HTRAG) mechanism by the output Topic Indicator Token $T_p$. Based on the topic-specific knowledge provided by TS3M, HTRAG can further enhance the generation procedure.

During the training, the total training objectives are defined as $\mathcal{L}_{Total}$:

\begin{equation}
\mathcal{L}_{Total} = \lambda_1 \mathcal{L}_{NSP} + \lambda_2 \mathcal{L}_{TH},
\end{equation}
where $\lambda_1$ and $\lambda_2$ are weighting factors. As for iterative fine-tuning, a higher weight might be placed on $\mathcal{L}_{NSP}$ in the beginning to establish foundational language understanding capabilities, with an increasing emphasis on $\mathcal{L}_{TH}$ as the model becomes more proficient at handling basic language structures.

The NSP loss, denoted as $\mathcal{L}_{NSP}$, is important for training the LLM to predict the logical continuation of a text sequence, which is implemented using a binary cross-entropy loss function:

\begin{equation}
\mathcal{L}_{NSP} = -\left[\mathbf{y} \log(\mathbf{\hat{y}}) + (1-y) \log(1-\mathbf{\hat{y}})\right],
\end{equation}
where $y$ is the ground truth label and $\mathbf{\hat{y}}$ is the model's predicted probability.

The Topic Hierarchy Loss $\mathcal{L}_{TH}$ ensures that the model's output captures the topic-related information as encoded in the hierarchical graph of topics. It is formulated to penalize the distance between the model-generated topic indicator token and the expected topic embeddings that follow the hierarchical relationships. The Topic Hierarchy Loss is given by:

\begin{equation}
\mathcal{L}_{TH} = e^{-d_{\text{graph}}(T, \hat{T})} \cdot \left(-\sum_{i=1}^{P} k_{i,T} \log(\hat{k}_{i,\hat{T}})\right),
\end{equation}
where $T$ is the true topic associated with the input, and $d_{\text{graph}}$ represents the shortest path distance in the hierarchical graph between the predicted topic $\hat{T}$ and $T$.


\subsection{HTRAG}

HTRAG is proposed to enhance the generation capabilities of the instruction-tuned model by integrating retrieved information during the inference. It aims to provide the model with relevant and context-specific knowledge, to improve the accuracy and information richness of generated responses.

HTRAG operates through a multi-step process that hierarchically retrieves and integrates topic-relevant information from an external knowledge base, which includes food encyclopedias, recipe databases, nutritional information, and culinary techniques. The knowledge base is preprocessed and indexed by FAISS~\cite{10.1162/tacl_a_00530} to facilitate efficient retrieval based on the input instructions. The gte-Qwen1.5-7B-instruct is implemented as the embedding model for retrieval.

The first step in HTRAG involves passing the input instruction tokens $X_q$ through the Topic-based Selective State Space Model (TS3M) to obtain the topic indicator token $\mathbf{K}$. Next, the topic indicator token obtained from TS3M is used to retrieve relevant information from the external knowledge base. The retrieval process employs a hierarchical approach that considers both the overall context of the input instructions and the specific topics identified by TS3M.

The retrieval process begins by matching the $\mathbf{K}$ against the indexed knowledge base $\mathbf{D}$ while calculating the similarity between the input queries and the content corresponding to the index using cosine distance. This initial retrieval step identifies a subset of relevant documents or passages that are semantically related to the input instructions. The retrieved information is then further processed and filtered based on the specific topics and their importance within the instructions.


Once the relevant information is retrieved, HTRAG integrates it with the original instruction representation. The integrated representation contains both the original instruction context and the supplementary knowledge retrieved from the external source. The augmented representation, which incorporates the retrieved topic-relevant information, is then passed to the LLM $\phi_\omega$ for generating the final response $\mathbf{T}_o$:

\begin{equation}
\mathbf{T}o = \phi_\omega(\text{Concat}(\mathbf{X}_q, \mathbf{R})),
\end{equation}
where $\mathbf{R}$ represents the integrated information from the retrieved documents, and $\text{Concat}(\cdot)$ denotes the concatenation operation. By hierarchically retrieving and integrating topic-relevant information, HTRAG enables the model to produce more informative and comprehensive responses to food-related queries. The integration of external knowledge enhances the model's ability to handle a wide range of food-related topics and makes our approach perform better in addressing more complex and individualized queries in the food domain.
\begin{figure*}[t]
	\centering
	\includegraphics[width=0.98\textwidth]{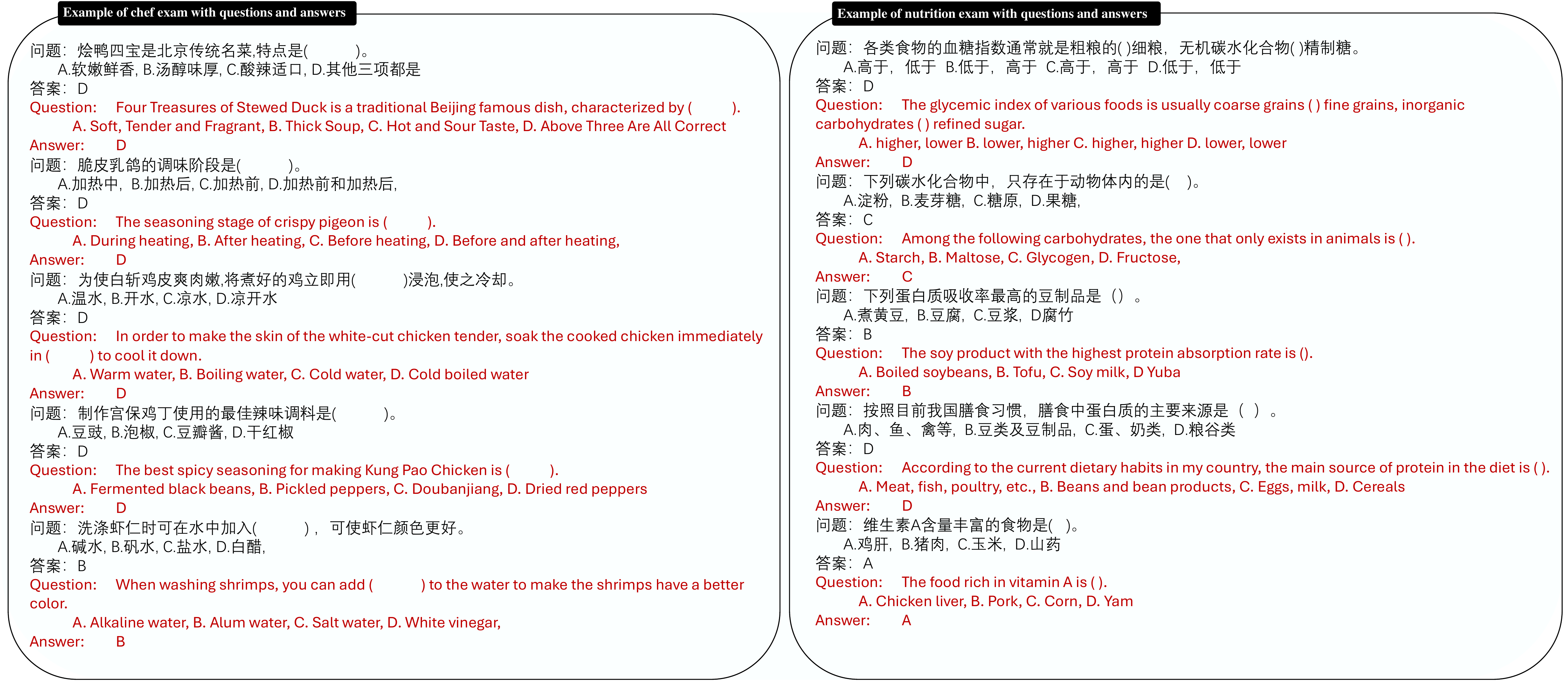}
	\caption{Examples of questions and answers in CDE benchmark with the corresponding English translation.}
	\label{fig:sample}
\end{figure*}

\begin{table*}[!htbp]
\caption{The zero-shot and few-shot (5-shot) performance of different LLMs on the CDE benchmark evaluated by the accuracy (\%).}
\centering
\setlength{\tabcolsep}{10pt}
\renewcommand{\arraystretch}{1.6}
\begin{tabular}{l c c c c c c }
\hline
 & \multicolumn{3}{c}{Zero-shot} & \multicolumn{3}{c}{Few-shot} \\
\cmidrule{2-4} \cmidrule{5-7}  & Chef Exam & Dietetic Exam & Total & Chef Exam & Dietetic Exam & Total \\
\hline
ChatGLM-6B\cite{chatglm1} & 49.1 & 42.9 & 43.8 & 44.5 & 47.4 & 47.0 \\
Mistral-7B\cite{mistral} & 30.9 & 42.4 & 40.7 & 44.5 & 48.2 & 47.6 \\
Vicuna-v1.5-7B\cite{vicuna2} & 42.5 & 36.7 & 36.5 & 40.9 & 45.6 & 44.9 \\
Baichuan2-7B\cite{yang2023baichuan} & 53.6 & 51.5 & 51.8 & 70.0 & 59.0 & 60.6 \\
CLLaMA2-7B\cite{cllama} & 39.7 & 40.9 &  39.9 & 41.9 & 35.4 & 41.0 \\
InternLM2-7B\cite{internlm2} & 55.8 & 68.2 &  57.6 & 63.0 & \textbf{75.4} & 64.8 \\
Qwen-7B\cite{qwen} & 49.0 & 42.7 & 48.0& 59.3 & 75.0 & 60.6 \\
\hline
Vicuna-v1.5-13B\cite{vicuna2} & 44.5 & 44.4 & 44.4 & 52.7 & 47.5 & 48.3 \\
CLLaMA2-13B\cite{cllama} & 48.2 & 53.6 & 52.7 & 44.0 & 45.3 & 48.2 \\
ChatGPT-3.5\cite{chatgpt} & 54.4 & 57.3 & 54.8 & 52.1 & 68.2 & 
54.5 \\
\hline
\textbf{FoodSky-7B (Ours)} & 62.0 & 60.1 & 61.7 & 62.2 & 62.8 & 62.4 \\
\textbf{FoodSky-13B (Ours)} & \textbf{67.2} & \textbf{66.4} & \textbf{67.1} & \textbf{69.5} & 70.9 & \textbf{69.7} \\
\hline
\end{tabular}
\label{tab:main}
\end{table*}

\section{Experiment}

\subsection{Experimental Setup}

\subsubsection{Implementation Details}
To investigate the impact of model size on performance, we experiment with two different backbone models: CLLaMA2-7B and CLLaMA2-13B. LLaMA-7B is a 7 billion-parameter model, while LLaMA-13B has 13 billion parameters, providing increased capacity for learning and generating responses. After the food-specific finetuning, we refer to the resulting models as FoodSky-7B for the 7B variant and FoodSky-13B for the 13B variant.

We train FoodSky-13B on a server with 8 NVIDIA A100 (80G) and train FoodSky-7B on a server with 8 NVIDIA V100 (32G), respectively. Our methods are trained using a Low-Rank Adaptation (LoRA) parameter-efficient tuning method~\cite{hu2021lora}. We use the transformers\footnote{https://huggingface.co/docs/transformers/} and the peft\footnote{https://github.com/huggingface/peft} libraries based on Pytorch. To make a trade-off between the training efficiency and performance, we implement fp12 precision with ZeRO-3\footnote{https://huggingface.co/docs/accelerate/usage\_guides/deepspeed} and gradient accumulation strategy based on DeepSpeed\footnote{https://github.com/microsoft/DeepSpeed}. We limit the length of a single response (including history) to 1,500. We use a dropout rate of 0.1, $10^{-5}$ and a cosine learning rate scheduler based on Adam~\cite{adam2014}. We evaluate the last convergent checkpoint as the final result. 

\subsubsection{Baseline Methods}

We compare FoodSky with representative state-of-the-art LLMs:

\begin{itemize}
\item ChatGLM2-6B \cite{chatglm2}: A 6B-parameter open bilingual LLM optimized for Chinese QA and dialogue. The model is trained for about 1 trillion tokens of Chinese and English corpus.
\item Mistral-7B \cite{mistral}: A 7B-parameter open LLM released by Mistral AI. The model is a carefully designed language model that provides both efficiency and high performance on various tasks to enable real-world applications.
\item InternLM2-7B \cite{internlm2}:A 7B-parameter open-source LLM tailored for practical scenarios. The model is trained on over 2 trillion high-quality pre-training corpora. It leverages trillions of high-quality tokens for training to establish a powerful knowledge base.
\item Vicuna-v1.5-7B and Vicuna-v1.5-13B \cite{vicuna1,vicuna2}: 7B-parameter and 13B-parameter open chatbots trained to replicate ChatGPT's behavior. The models are fine-tuned from Llama 2 with supervised instruction fine-tuning. The training data is around 125K conversations collected from ShareGPT.com.
\item Baichuan2-7B \cite{yang2023baichuan}: A 7B-parameter multilingual LLM released by Baichuan Intelligence. The model is trained on a high-quality corpus with 2.6 trillion tokens.
\item Qwen-7B \cite{qwen}: A 7B-parameter open-source LLM released by Alibaba Group, demonstrating superior performance in multiple downstream task. The model is pretrained for up to 3 trillion tokens of multilingual data with a wide coverage of domains.
\item ChatGPT-3.5 \cite{chatgpt}: The chatbot developed by OpenAI. The model is trained using reinforcement learning from human feedback and fine-tuned from GPT-3.5.
\item CLLaMA2-7B \cite{cllama}: A 7B-parameter Chinese pre-trained version of the open-source LLM LLaMA \cite{touvron2023llama}. A method to augment LLaMA with capabilities for understanding and generating Chinese text and its ability to follow instructions is utilized in pre-training and fine-tuning process.
\end{itemize}

\subsubsection{Evaluation}
To comprehensively evaluate different LLMs, we first employ the accuracy as the main metric to compare the performance on the CDE benchmark. Then, a range of automatic evaluation metrics are commonly used in natural language generation tasks. These include BLEU-1, BLEU-2, BLEU-3 and BLEU-4 \cite{papineni2002bleu}, which measure the n-gram overlap between the generated and reference answers; ROUGE-1, ROUGE-2, and ROUGE-L \cite{lin2004rouge}, which assess the unigram, bigram, and longest common subsequence overlap, respectively; and GLEU \cite{wu2016googles}, which is a variant of BLEU that correlates better with human judgments. Additionally, Distinct-1/2 \cite{li2016diversity} assesses the textual diversity of the generated answers by calculating the ratio of distinct unigrams and bigrams. We also use GPT-4 as the judge to subjective scores of fluent score, logic score, professional score, informative score on FoodQA.

We test the baseline models and FoodSky on the Chef and Dietetic Examinations (CDE) benchmark, Food Long Conversation (FoodLongConv) benchmark and Food Question and Answer (FoodQA) benchmark, respectively. In the CDE benchmark, we use the standard choice questions in the Chinese Chef Examination and the Dietetic Examination as our benchmarks to evaluate the performance of different LLMs in the food and nutrition domain. Specifically, we collect 628 choice questions from the Chinese Chef Examination and 111 choice questions from the Chinese Dietetic Examination to challenge LLMs in the setting of standard chef and dietetic exams\footnote{http://www.zyzgks.net/index\_15.html}. The accuracy of the CDE benchmark is also the main metric for evaluating LLMs since it is the most objective indicator that determines whether the answer of LLMs is correct or not.

To evaluate the conversation and question-answering capabilities of the LLMs, we build FoodLongConv benchmark and FoodQA benchmark. The FoodLongConv contains 22 essay questions that require long answers, extracted from the Chinese Chef Examination and the Chinese Dietetic Examination. We use the metrics mentioned above to evaluate the quality of LLMs answers on FoodLongConv. The FoodQA is a benchmark focusing on comparing different performances of the model's answers and contains 25 short answer questions and essay questions extracted from the Chinese Chef Examination and the Chinese Dietetic Examination. Some examples are shown in the Fig.~\ref{fig:sample}. We use GPT-4 as the judge to score LLMs' answers with respect to fluency, logic correctness, professionality and information density on FoodQA.

\subsection{Comparison with Baseline Methods}

\subsubsection{Comparison on the CDE benchmark}

\begin{table*}[!htbp]
\caption{Performance comparison of different models on different topic subcategories (\%).}
\centering
\setlength{\tabcolsep}{4.5pt}
\renewcommand{\arraystretch}{1.5}
\begin{tabular}{lccccccc}
\hline
& Dietary Science & Food Flavor Profiles & Food Safety Measures & Food Recipes & Healthy Eating Principles & Average \\
\hline
Baichuan2-7B\cite{yang2023baichuan} & 56.1 & 59.8 & 46.8 & 49.0 & 60.1 & 51.8 \\
CLLaMA2-7B\cite{cllama} & 40.8 & 45.8 & 33.0 & 37.9 & 42.4 & 39.9 \\
InternLM2-7B\cite{internlm2} & 60.5 & 58.9 & 61.7 & 51.9 & 70.0 & 57.6 \\
Qwen-7B\cite{qwen} & 50.3 & 57.0 & 45.7 & 45.8 & 43.9 & 48.0 \\
ChatGPT-3.5\cite{chatgpt} & 66.9 & 56.1 & 57.4 & 48.7 & 59.1 & 54.8 \\
\hline
\textbf{FoodSky-7B (Ours)} & \textbf{70.1} & 62.6 & 62.8 & 56.3 & 62.1 & 61.7 \\
\textbf{FoodSky-13B (Ours)} & 68.3 & \textbf{68.2} & \textbf{67.0} & \textbf{65.6} & \textbf{73.1} & \textbf{67.1} \\
\hline
\end{tabular}
\label{tab:subtopic}
\end{table*}


\begin{table*}[!htbp]
\caption{Performance comparison of different models on FoodLongConv benchmark (\%).}
\centering
\setlength{\tabcolsep}{5pt}
\renewcommand{\arraystretch}{1.5}
\begin{tabular}{lcccccccccc}
\hline
& BLEU-1 & BLEU-2 & BLEU-3 & BLEU-4 & GLEU & ROUGE-1 & ROUGE-2 & ROUGE-L & Distinct-1 & Distinct-2\\
\hline
Baichuan2-7B\cite{yang2023baichuan} &  23.19 &  10.24 &  6.58 &  3.36  &  14.64  & 29.78 &  7.84 & 23.31 &  50.79 & 81.32  \\
CLLaMA2-7B\cite{cllama} &  12.59 &  5.50 &  2.38 &  0.97 &  0.97 &  27.18 &  8.15 &  21.55 & \textbf{67.99}  & \textbf{90.83} \\
InternLM2-7B\cite{internlm2} &  16.29  &  5.91  &  3.22 &  1.08  &  8.73  &  24.90  &  3.79  &  19.57 &  33.41 & 64.37 \\
Qwen-7B\cite{qwen} & 19.18  & 9.01  & 2.37  &  1.83  &  12.33  & 25.73  & 5.25  & 20.71  &  60.64 &  87.11 \\
ChatGPT-3.5\cite{chatgpt} & 23.54 &  10.66 & 4.44  &  1.43 &  13.12  &  28.67 &  6.53  & 22.93 & 46.46 & 76.36 \\
\hline
\textbf{FoodSky-7B (Ours)} & 22.95  & 13.53 & 9.72 & 7.34 & 16.18 & 28.79 & 8.00 & 22.44  & 60.83 & 87.32 \\
\textbf{FoodSky-13B (Ours)} & \textbf{24.84} &  \textbf{15.27}  &  \textbf{11.01}  &  \textbf{8.89} & \textbf{18.26}  & \textbf{34.50}  &   \textbf{14.08} & \textbf{28.78}  & 57.81 & 82.54 \\
\hline
\end{tabular}
\label{tab:nlp}
\end{table*}

Table~\ref{tab:main} presents the zero-shot and few-shot (5-shot) performance of different LLMs on the CDE benchmarks, where the accuracy on questions from the Chef Exam and Dietetic Exam and mean accuracy are reported respectively.

In the zero-shot setting, FoodSky-7B and FoodSky-13B models achieve the highest accuracy on both Chef and Dietetic exams, outperforming all other baseline models. Specifically, FoodSky-7B achieves 62.0\% accuracy on the Chef Exam and 60.1\% accuracy on the Dietetic Exam, resulting in a total accuracy of 60.2\%. FoodSky-13B further improves the performance, achieving 67.2\% accuracy on the Chef Exam and 66.4\% accuracy on the Dietetic Exam, with a total accuracy of 67.1\%. Among the baseline models, InternLM2-7B and ChatGPT-3.5 perform the best, with 57.6\% and 54.8\% accuracy respectively, followed by Baichuan2-7B (51.8\%) and CLLaMA2-13B (52.7\%).

In the few-shot setting, all models show improved performance compared to the zero-shot setting, benefiting from the additional context provided by few-shot examples. FoodSky-13B maintains its lead, achieving a total accuracy of 69.7\%, with 69.5\% on the Chef Exam and 70.9\% on the Dietetic Exam. FoodSky-7B closely follows with a total accuracy of 61.8\%. InternLM2-7B and ChatGPT-3.5 also perform well, with total accuracy rates of 64.8\% and 54.5\%, respectively. Baichuan2-7B achieves a total accuracy of 60.6\%, while CLLaMA2-13B obtains total accuracy of 48.2\%, respectively.

The superior performance of FoodSky models can be attributed to their specialized training on a large-scale food and nutrition corpus, which enables them to better understand and reason about the domain-specific concepts and principles covered in the Chef and Dietetic Examinations. The results demonstrate that proposed FoodSky has the ability to pass the both Chinese Chef Examination and the Dietetic Examination, providing an available domain-specific LLM for the food and nutrition domain.

\subsubsection{Comparison on Different Topic Categories}

We use the CDE, which contains choice questions from Chef and the Dietetic examinations to evaluate the performance of different methods on different topic categories. Specifically, we divide the topics in the CDE benchmark into five main categories: Dietary Science, Food Flavor Profiles, Food Safety Measures, Food Recipes, and Healthy Eating Principles. 

As shown in Table~\ref{tab:subtopic}, FoodSky-13B achieves the highest scores in every topic subcategory with significant results. Specifically, FoodSky-13B achieves an accuracy of 68.3\% in ``Dietary Science" and an accuracy of 68.2\% in ``Food Flavor Profiles", demonstrating a capable understanding of nutritional queries, which is fundamental to the dietetic examinations. FoodSky-7B also shows strong performance with an accuracy of 70.1\% and 62.6\%, respectively. The second best model, ChatGPT-3.5, follows with an accuracy of 66.9\% in the ``Dietary Science" and an accuracy of 59.1\% in ``Healthy Eating Principles".

With respect to other topic categories, FoodSky-13B also outperforms all other models with an accuracy of 67.0\% in ``Food Safety Measures", 65.6\% in ``Food Recipes" and 73.1\% in ``Healthy Eating Principles". Its overall average accuracy is 67.1\%. FoodSky-7B achieves notable performance with 62.8\% in ``Food Safety Measures", 56.3\% in ``Food Recipes" and 62.1\% in ``Healthy Eating Principles", averaging 61.7\%.

Overall, the average performance across all categories for FoodSky-13B stands at 62.9\%. The comparative analysis of different topic subcategories shows that FoodSky can generate more precise responses across diverse topics within the food and nutrition domain, compared other general models which show varied performance across these topic categories.

\begin{figure*}[t]
	\centering
	\includegraphics[width=0.9\textwidth]{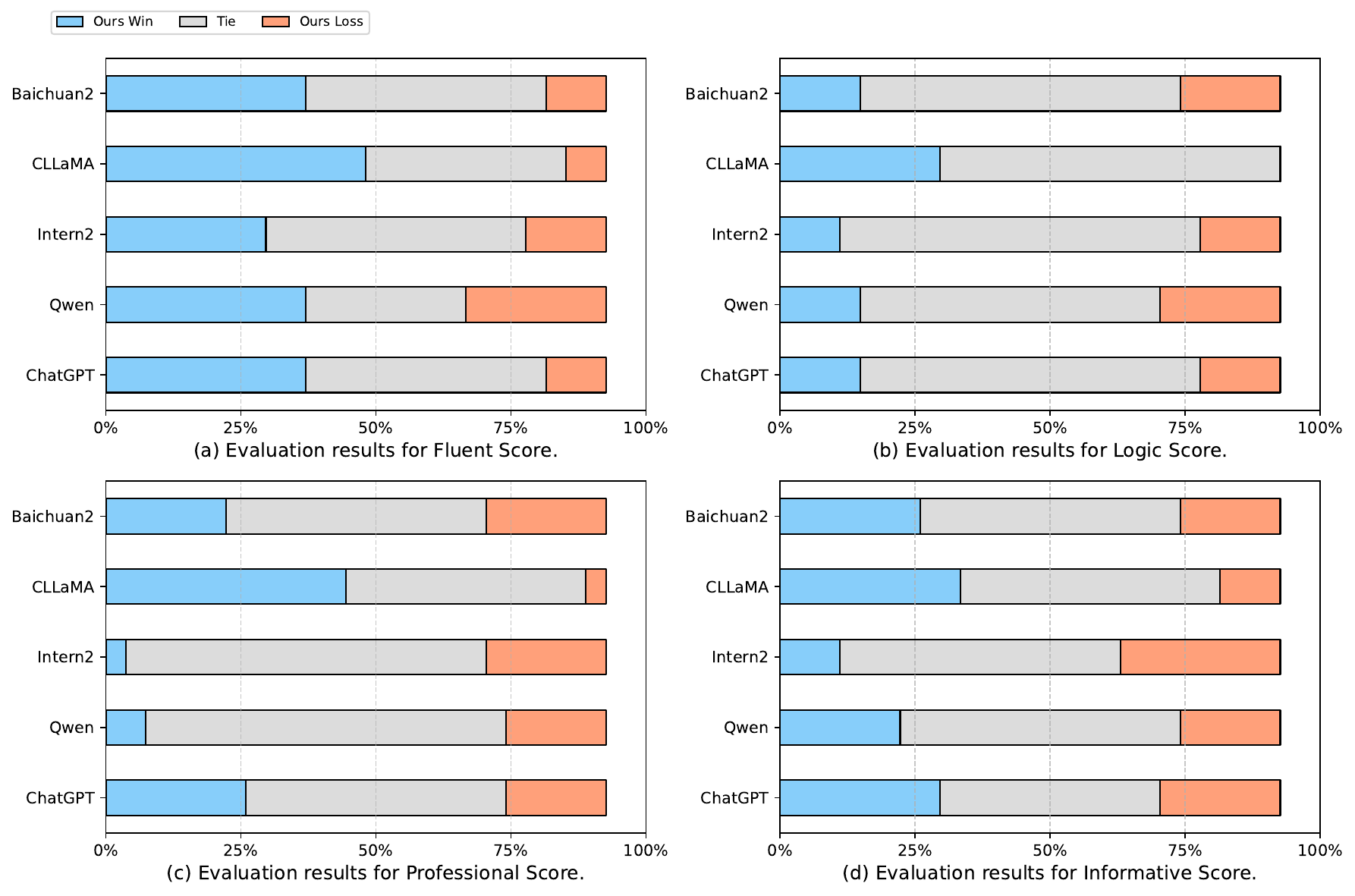}
	\caption{Performance comparison of our FoodSky-7B with different baseline models on FoodQA benchmark evaluated by GPT-4. GPT-4 gives the Fluent Score, Logic Score, Professional Score and Informative Score based on the answers of the models, and judges which model wins on each score.}
	\label{fig:pk}
\end{figure*}

\begin{table*}[!t]
  \centering
  \caption{Abalation studies on model structures (\%). The baseline model is the FoodSky-7B trained on the FoodEarth-680K dataset.}
  \centering
    \setlength{\tabcolsep}{2.2pt}
    \renewcommand{\arraystretch}{1.5}
  \begin{tabular}{cccccccccccp{1.0cm}<{\centering}p{1.1cm}<{\centering}p{1.4cm}<{\centering}p{1.2cm}<{\centering}}
    \toprule
    \multicolumn{1}{c}{\multirow{2}[4]{*}{Model}} & \multicolumn{2}{c}{Methods}  & \multicolumn{3}{c}{CDE} & \multicolumn{5}{c}{FoodLongConv} & \multicolumn{4}{c}{FoodQA}\\ 
\cmidrule{2-6}\cmidrule{7-11}\cmidrule{12-15} & TS3M & HTRAG & Chef & Dietetic & Total  & BLEU-1  &  BLEU-4 &  GLEU & ROUGE-1  & ROUGE-L & Fluent & Logic & Professional & Informative\\
    \midrule
    \multirow{3.5}[4]{*}{FoodSky-7B} & &  & 53.4 & 53.6 &  53.5 & 16.41 & 6.62 & 13.51 & 26.07 & 20.30 & 86.25 & 73.80 & 72.20 & 74.00 \\
      &    \checkmark &  & 56.8 & 60.0 & 57.2 & 18.40 & 6.09 & 14.09 & 26.57 & 21.09 & 89.00 & 77.20 & 77.20 & 81.80 \\
       &   &  \checkmark  & 55.8 & 59.1 & 56.3 & 18.89 & \textbf{7.18} & 14.42 & 26.21 & \textbf{21.16} & 85.23 & 75.68 & 75.00 & 74.77 \\
       & \checkmark     & \checkmark     & \textbf{60.5} & \textbf{60.7} & \textbf{60.5} & \textbf{19.78} & 6.63 & \textbf{14.50} & \textbf{27.42} & 21.15 & \textbf{89.32} & \textbf{82.27} & \textbf{81.82} & \textbf{84.55} \\
    \bottomrule
    \end{tabular}
  \label{tab:ablation1}%
\end{table*}%

\begin{figure*}[t]
	\centering
	\includegraphics[width=0.9\textwidth]{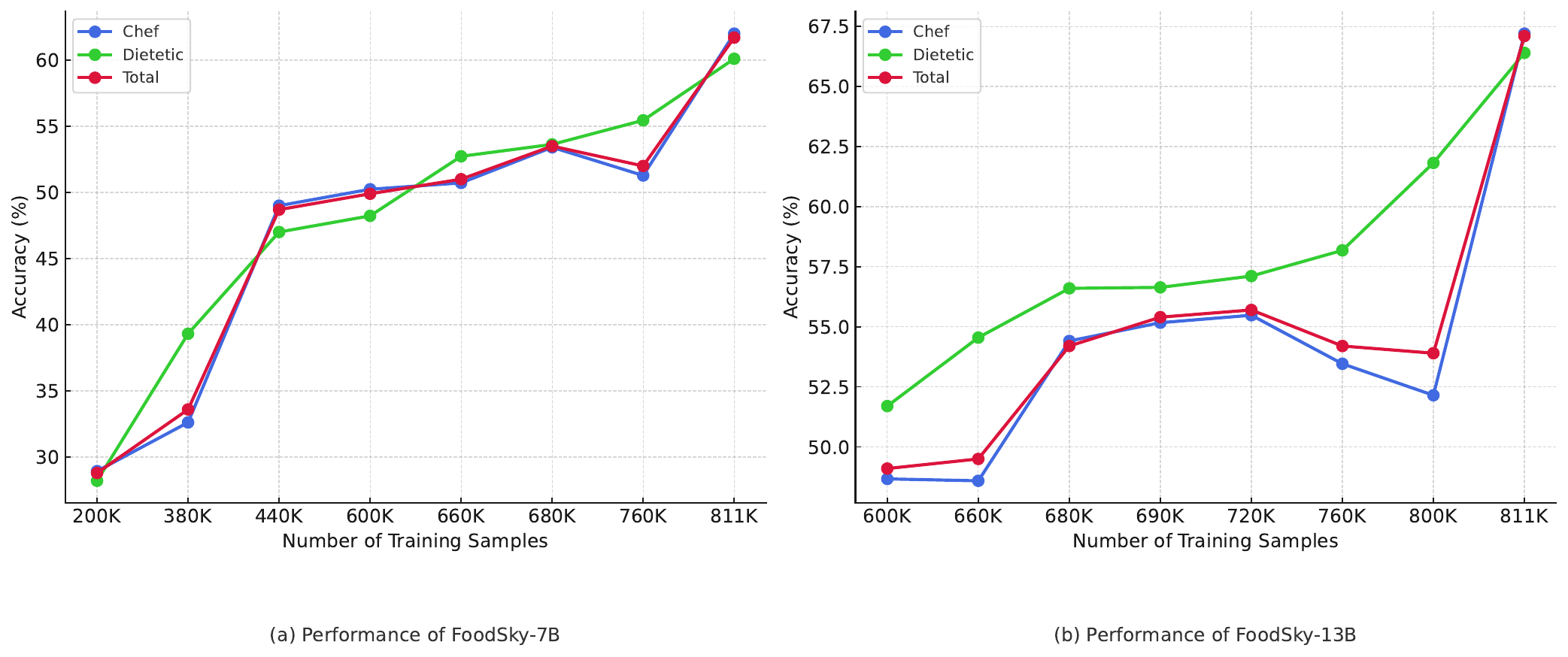}
	\caption{Ablation study on instruction numbers in the training set. The results are the accuracy on the CDE benchmark.}
	\label{fig:ablation2}
\end{figure*}

\subsubsection{Comparison on the Question Answering Task}
We use the FoodQA benchmark to evaluate different models on a question-answering task related to the food and nutrition domain. Experimental results are detailed in Table~\ref{tab:nlp} and analyzed using NLP metrics such as BLEU, GLEU, ROUGE, and Distinct.

FoodSky-13B outperforms all models across most metrics, showing its robustness in generating detailed and relevant responses. It achieves the best in BLEU-1 (24.84\%), BLEU-2 (15.27\%), BLEU-3 (11.01\%), BLEU-4 (8.89\%), GLEU (18.26\%), ROUGE-1 (34.50\%) and ROUGE-2 (14.08\%), and ROUGE-L (28.78\%), indicating the accuracy and linguistic richness in its responses. FoodSky-7B also shows capable performance, especially in BLEU-4 (7.34\%) and GLEU (16.18\%), showing its effectiveness in producing varied and precise language. 

As for Distinct-1 and Distinct-2, compared with FoodSky, the baseline CLLaMA2-7B with lower accuracy achieves the highest Distinct scores since it does not have the knowledge to actually answer several questions and outputs more uncontrolled responses. With the improvement of professionalism and answer rigor, the answers tend to follow a certain correct logic, therefore influencing the output diversity of answers. Compared with the strong baseline InternLM2-7B and Qwen-7B, our FoodSky maintains a better balance between high-quality responses and diversity of responses. Baichuan2-7B displays strong performance in ROUGE metrics, with the significant ROUGE-1 score (29.78\%) and ROUGE-L score (23.31\%), highlighting its effectiveness in capturing key content from the context.

Overall, the results from the FoodQA benchmark confirm the superiority of FoodSky in understanding and responding to complex questions in the food and nutrition domain, still setting it apart from other general models that show varied performance across these metrics.

\begin{figure*}[t]
	\centering
	\includegraphics[width=0.9\textwidth]{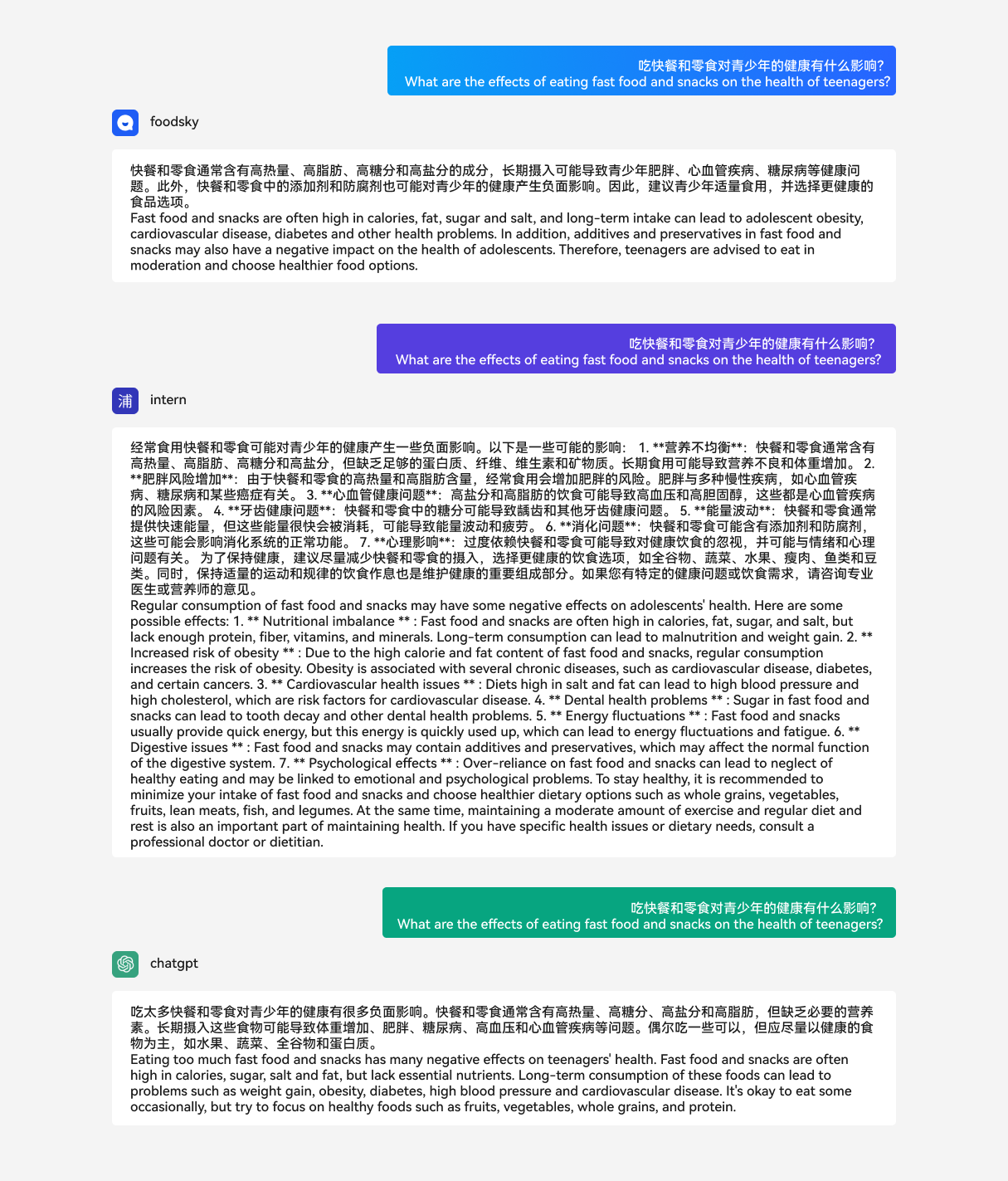}
	\caption{Qualitative experimental results of question answering visualization by FoodSky, InternLM2 and ChatGPT-3.5. FoodSky. While the outputs from Intern and ChatGPT are relatively redundant and lack focus, the answers from our FoodSky are more accurate and concise.}
	\label{fig:teenager}
\end{figure*}


\subsubsection{Evaluation with GPT-4 Score}
In addition to the objective evaluation metrics, we also conducted a subjective evaluation using GPT-4 as the judge to assess the quality of the generated responses from different models. We used questions from the FoodQA benchmark and asked each model to generate responses. GPT-4 then evaluated the responses based on four criteria: fluent score, logic score, professional score and informative score.

Fig.~\ref{fig:pk} illustrates the performance comparison of different models based on GPT-4 assessment. FoodSky-7B achieves the highest score in fluency with a fluent score of 40. Among the baseline models, InternLM2-7B (Intern2) and Qwen-7B (Qwen) demonstrate strong performance, with scores ranging from 20\% to 35\% across the three criteria of logic score, professional score and informative score. ChatGPT-3.5 (ChatGPT) and Baichuan2-7B (Baichuan) also show competitive results in these three criteria, with scores between 20\% and 30\%. The baseline CLLaMA2-7B (CLLaMA), achieves lower scores, ranging from 0\% to 20\% in the four criteria.

The subjective evaluation shows that the baseline models lag behind in answering multiple-choice questions and can still generate reasonable answers judged by GPT-4. It also confirms the reliable performance of FoodSky models in generating high-quality, professional, and informative responses in the food and nutrition domain. However, the Q\&A ability of the proposed FoodSky can be further improved by reinforcement learning with human feedback.

\begin{figure*}[t]
	\centering
	\includegraphics[width=0.7\textwidth]{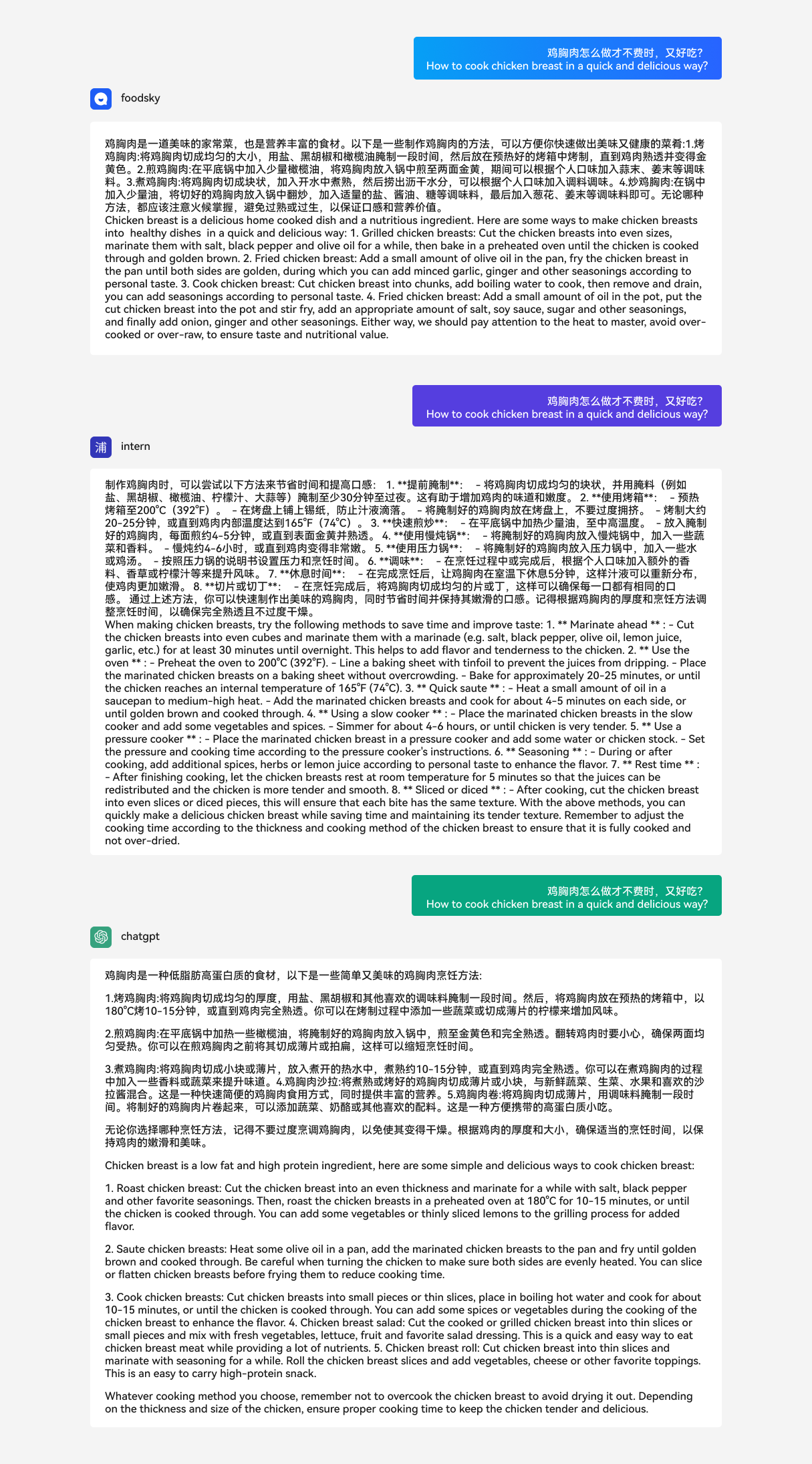}
	\caption{More qualitative experimental results of question answering visualization by FoodSky, InternLM2 and ChatGPT-3.5. The answers from our FoodSky are more informative than those from Intern and ChatGPT.}
	\label{fig:chef}
\end{figure*}

\subsection{Ablation Studies}

To investigate the impact of different components in FoodSky, we conducted ablation studies by removing key components and evaluating the performance on the CDE, FoodLongConv and FoodQA benchmarks. In order to swiftly verify the validity of the proposed methods at the early stage of data construction, we use FoodEarth-680K, a high-quality version of FoodEarth with 680K instruction data to train our FoodSky. Table~\ref{tab:ablation1} presents the results of ablation studies on model structure.

The baseline FoodSky-7B model, without any additional components, obtains an average accuracy of 53.5\% in the CDE benchmark. Incorporating TS3M improved the average accuracy to 57.2\%, an improvement of 3.7\%, demonstrating its importance in enhancing topic understanding. Further, integrating HTRAG component resulted in a score of 56.3\%, with an improvement of 2.8\%, suggesting its effect in providing contextually relevant answers by leveraging external knowledge.

The most notable improvements were observed when both TS3M and RAG were simultaneously employed, pushing the average score to 60.5\%. This configuration also excelled in the FoodQA dataset, achieving the highest metrics across BLEU-1 (19.78\%), BLEU-4 (6.63\%), GLEU (14.50\%), and ROUGE-L (21.15\%). These scores confirm that the combination of TS3M and RAG can produce the most effective performance in generating accurate and relevant responses. The ablation study highlights the key role of TS3M and RAG in improving the performance of FoodSky-7B across different linguistic and semantic challenges, highlighting their synergies in improving model accuracy and response quality.





\subsection{Evaluation on Different Numbers of Training Data}
To assess the impact of training set size on the performance of FoodSky, we conducted ablation studies where both FoodSky-7B and FoodSky-13B were trained with varied numbers of instructions from the FoodEarth-680K dataset. The results in Fig.~\ref{fig:ablation2} demonstrate how changes in data volume influence model accuracy and linguistic metrics.

For FoodSky-7B, increasing the training set size lead to consistent improvements in performance on the CDE benchmark. Starting from 200K samples, the total accuracy on the CDE benchmark rose steadily from 28.8\% to 61.7\%. The FoodSky-13B model exhibits a more significant improvement with the expansion of the dataset. Notably, when trained with 600K samples, the model already achieves an accuracy of 48.67\% on the chef exam in the CDE benchmark, which jumps to 67.2\% by 811K samples. The total accuracy follows a similar trajectory, starting at 49.1\% and reaching 67.1\%.

It is worth noting that, when we increase the data sample to 780K, the accuracy of both FoodSky-7B and FoodSky-13B decreases in Fig.~\ref{fig:ablation2}. This shows that simply increasing the amount of data does not necessarily improve the performance of the model. The amount of data we initially supplied from 680K to 800K, was provided by directly adding data extracted from public teaching datasets that were not double-checked for data quality. This suggests that data quality is also important for improving the accuracy of the model in chef and diet tests, even more important than pure data volume. 
We removed all the low-quality and duplicate data when we built the final version of FoodEarth, resulting in 811K high-quality samples.

The ablation study highlights the substantial benefits of scaling up the training data. FoodSky-13B, with its larger capacity, effectively leverages additional data to surpass the performance of FoodSky-7B. With the full training set of 811K instructions, FoodSky-13B achieves notable accuracy rates: 67.2\% on the chef exam, 66.4\% on the dietetic exam, and a total accuracy of 67.1\% on the CDE benchmark. This demonstrates that the model can generate more accurate and informative responses as training data increases.




\subsection{Qualitative Results}


As illustrated in Fig.~\ref{fig:chef}, the qualitative experimental results in the scenario of recipe recommendation highlight both strengths and weaknesses in the predictive capability of different LLMs. In the first example, FoodSky correctly recognizes the key information that describes ways to make chicken breast more delicious. This capability reflects a strong understanding of the fundamental principles of culinary arts. However, in Fig.~\ref{fig:chef}, the baseline models Intern and ChatGPT demonstrate inaccuracies in capturing detailed culinary nuances. Specifically, they struggle with the richness of the cuisines. The Intern even overlooks the key requirement of saving time and effort. These mistakes suggest a gap in the model's ability to handle complex cooking instructions and ingredient interactions effectively.

Three more qualitative results in Fig.~\ref{fig:teenager}, which shows the examples of dietary education for adolescents, reveal the different focus and depth of analysis of different methods. FoodSky delivers the most in-depth exploration of the topic, focusing on the direct health impacts of fast food snacks, promoting moderation consumption for teenagers. The answers of Intern are lengthy as it offers too detailed health effects and extensive dietary advice. ChatGPT presents a balanced view that integrates moderate consumption of fast food with healthier dietary habits. Overall, FoodSky is capable of quickly identifying the immediate health risks associated with high-calorie and high-fat diets and providing actionable recommendations, which are critical to promoting healthier eating habits in teenagers.



\section{Conclusion}
In summary, FoodSky stands as a robust food-specific LLM that provides new directions for future research and applications in the food field. To ensure the success of FoodSky, we first provided the basis for model training by building a large-scale, high-quality food corpus that contains various types of food-related instruction data. Second, we proposed the Topic-Based Selective State Space Model (TS3M) and Hierarchical Topic Retrieval-Enhanced Generation (HTRAG) to enhance FoodSky in processing and generating food-related content. Through extensive experiments, FoodSky has demonstrated significant capabilities in understanding and generating food-related content, surpassing existing general-purpose LLMs in both chef and nutrition exams. In the future, FoodSky will have broad prospects in several directions. First, by combining FoodSky with reinforcement learning enhanced by user feedback, the model can continually refine its understanding and generation capabilities. Second, FoodSky can also be extended as a Multimodal Large Language Model (MLLM). For example, recipe suggestions can be obtained through pictures of ingredients, and future weight changes can be predicted through nutritional analysis of dishes. We will also further introduce more data from the food industry, to create an LLM specifically for the food industry. This will enable intelligent transformation and upgrading in the food sector, with implications for key areas such as food design, food safety, and supply chain management.

{\appendix 

We list the data sources in detail in the appendix, including authoritative public accounts and websites, knowledge graph data, and public Chinese teaching datasets, and attach question-answering images of three different LLMs to demonstrate the question-answering comparison experiment between Foodsky and other models.

\subsection{Data Sources Details}
Table~\ref{tab:data-source-public-account} includes a detailed list of the data sources and descriptions of authoritative public accounts and websites used for this study. The table provides the name of each data source, a brief description of the specific content, and the corresponding citation source. These data sources cover public accounts related to nutrition, clinical nutrition services, medical health information, food culture, etc., as well as academic resources such as China National Knowledge Infrastructure (CNKI), providing rich data support for the study.

\begin{table*}[htbp]
\caption{Detailed data source information of authoritative public account and authoritative website.}
\label{tab:data-source-public-account}
\centering
\renewcommand{\arraystretch}{1.5}
\begin{tabular}{l p{10cm}<{\centering} p{2cm}<{\centering}}
\hline
Name & Description & Source Citation \\
\hline
China Good Nutrition & It is a public account operated by the Chinese Nutrition Society or related nutrition professional organizations. It is dedicated to popularizing nutrition knowledge, providing scientific dietary guidance, and sharing healthy lifestyle information, and may involve The latest nutritional research results. This public account aims to help the public understand how to improve health, prevent disease, and enhance overall quality of life through proper diet. Over 2000 articles were collected from January to February 2024. [Accessed 28-2-2024] & Authoritative public account: China Good Nutrition
\\
Clinical Nutrition Network & Co-founded by over 1500 nutritionists and physicians from over 700 public hospitals nationwide, it provides clinical nutrition services and includes several thousand posts from January to February 2024. [Accessed 28-2-2024]  & Authoritative public account: Clinical Nutrition Network
\\
Dr. Lilac & It is a popular health and medical information public account in China. it provides the public with medical and health knowledge, disease prevention, healthy lifestyle suggestions, and the latest medical information. Medical professionals usually run this public account which aims to increase public health awareness and help people better understand and manage their health. The dataset includes several thousand posts from January to February 2024. [Accessed 28-2-2024] & Authoritative public account: Dr. Lilac
\\
Chestnut-eating Pie & It is a WeChat public account focusing on food, nutrition and food culture. It provides a variety of relevant content for lovers of healthy eating and cooking, including but not limited to recipe sharing, ingredient introduction, nutritional knowledge, dietary tips and cooking techniques. Public accounts may also involve food safety, diet trend analysis, and promotion of healthy eating and lifestyle. The dataset includes several thousand posts from January to February 2024. [Accessed 28-2-2024] & Authoritative public account: Chestnut-eating Pie
\\
CNKI & CNKI (China National Knowledge Infrastructure) is one of the largest academic information resource providers in China. It provides a wide range of academic resources, including journal articles, theses, conference papers, standards, patents, books, etc. Through the keywords of disease and diet, more than 1,400 related papers were initially downloaded. [Accessed 31-4-2024] &  Articles downloaded in April 2024, containing several thousand posts. \\
\hline
\end{tabular}
\end{table*}

Table~\ref{tab:data-source-Knowledge-graph-data} provides the main sources of knowledge graph data and their descriptions. The knowledge graph data covers food nutrition data and nutrition knowledge data collected from multiple authoritative websites from 2021 to 2022. Through multiple rounds of data cleaning and construction, a detailed knowledge graph was finally formed, providing solid data support for the research.

\begin{table*}[htbp]
\caption{Detailed data source information of knowledge graph data.}
\label{tab:data-source-Knowledge-graph-data}
\renewcommand{\arraystretch}{1.5}
\centering
\begin{tabular}{l p{7cm}<{\centering} p{3cm}<{\centering}}
\hline
Name & Description & Source Citation \\
\hline
 Food nutrition data & The knowledge graph collected data from multiple authoritative websites from 2021 to 2022. Through multiple rounds of data set cleaning and construction for three months by numerous people, it completed the construction of a knowledge graph with more than 283,000 relationship nodes and a total number of more than 80,000 nodes. & Institute of Nutrition and Health, Chinese Center for Disease Control and Prevention, Chinese Food Composition List, China Public Health Science Data Center, Chinese Nutrition Society, USDA Food Nutrition Facts Database \\
 Nutrition knowledge data & The Nutrition Knowledge Knowledge Graph collected data from multiple authoritative websites from 2021 to 2022. Through multiple rounds of cleaning and construction of the data set by multiple people, it completed the construction of 2,045 relationship nodes and a total number of more than 35,000 nodes, based on the Food Nutrition Knowledge Graph for additional connections. &  U.S. Food and Drug Administration (FDA), European Food Safety Authority (EFSA), China State Food and Drug Administration (CFDA)\\
\hline
\end{tabular}
\end{table*}

Table~\ref{tab:data-source-of-public-Chinese-instruction-dataset} contains a table listing the main sources of public Chinese command datasets used for this study. These datasets were collected from multiple authoritative platforms, including GitHub, Hugging Face, and other open-source projects. A large amount of food-related conversation data was screened by keywords, providing important data support for the study.

\begin{table*}[htbp]
\caption{Detailed data source information of public Chinese instruction dataset.}
\label{tab:data-source-of-public-Chinese-instruction-dataset}
\renewcommand{\arraystretch}{1.5}
\centering
\begin{tabular}{l p{7cm}<{\centering} p{3cm}<{\centering}}
\hline
Name & Description & Source Citation \\
\hline
 ChatGPT Chinese corpus & The data set is a general scene Chinese conversation data set published in Git Hub, and 81,475 pairs of food-related data were filtered out through the tags set by the author. & Dataset downloaded in April 2024. [Online; accessed 30-4-2024], https://github.com/PlexPt/chatgpt-corpus \\ 
 COIG & The COIG data set is an open-source general scene conversation data set released by the Beijing Academy of Artificial Intelligence on huggingface. There are 898 pairs of food-related data sets filtered by keywords. & Dataset downloaded in April 2024. [Online; accessed 30-4-2024], https://huggingface.co/datasets/BAAI/COIG \\ 
 The Chinese dataset of Instruction Tuning with GPT-4 & This data set is used in the paper published by Microsoft Research: INSTRUCTION TUNING WITH GPT-4. This data set is a Chinese and English conversation data set in a general scenario generated by ChatGPT-4. We use food keywords to find 8828 pairs of related datasets in the Chinese conversation data set & Dataset downloaded in April 2024, [Online; accessed 30-4-2024], https://github.com/Instruction-Tuning-with-GPT-4/GPT-4-LLM/tree/main/data \\
 The Chinese dataset of RefGPT & RefGPT data set was jointly constructed by NLP practitioners from Shanghai Jiao Tong University, Hong Kong Polytechnic University and other institutions. Through fine-tuned GPT models, multiple rounds of dialogue data sets based on fact-based documents were generated. We filtered out more than 16,000 pieces of food-related conversation data from the Chinese data set. & Dataset downloaded in April 2024, [Online; accessed 30-4-2024],  https://github.com/ziliwangnlp/RefGPT \\ 
 The Chinese dataset of BELLE & This data set is a 3.5 million Chinese data set generated by ChatGPT3.5 provided by an open source project with more than 7,600 stars on GitHub. Through keyword search, we initially found more than 100,000 food-related data sets. & Dataset downloaded in April 2024. [Online; accessed 30-4-2024],  https://github.com/LianjiaTech/BELLE \\ 
 MOSS Chinese dataset  & This data set is an open-source conversation data set of Fudan University used to train MOSS-002, with a total of more than 600,000 items. We initially screened out 70,000 food-related questions and answers through keywords. & Dataset downloaded in April 2024. [Online; accessed 30-4-2024], https://huggingface.co/datasets/fnlp/moss-002-sft-data \\ 
\hline
\end{tabular}
\end{table*}

\subsection{More Qualitative Experimental Results}
We provide more qualitative experimental results in this appendix, including 8 figures showing the responses of three different language models in different groups. The Fig~\ref{fig:kidQA},~\ref{fig:BodyBuilderQA},~\ref{fig:SeniorPeopleQA},~\ref{fig:NutritionWorkerQA},~\ref{fig:diseasePolulationQA},~\ref{fig:BrainWorkerQA},~\ref{fig:ManualWorkerQA} and \ref{fig: PregnantQA} include questions and answers focus on issues for infants, fitness people, the elderly, nutrition and health workers, chronic disease patients, mental workers, manual workers, and pregnant women. Each figure records the performance of different language models when answering the same question. Our answers are concise and to the point, solving the core problems of the user groups.

\begin{figure*}[t]
	\centering
	\includegraphics[width=0.7\textwidth]{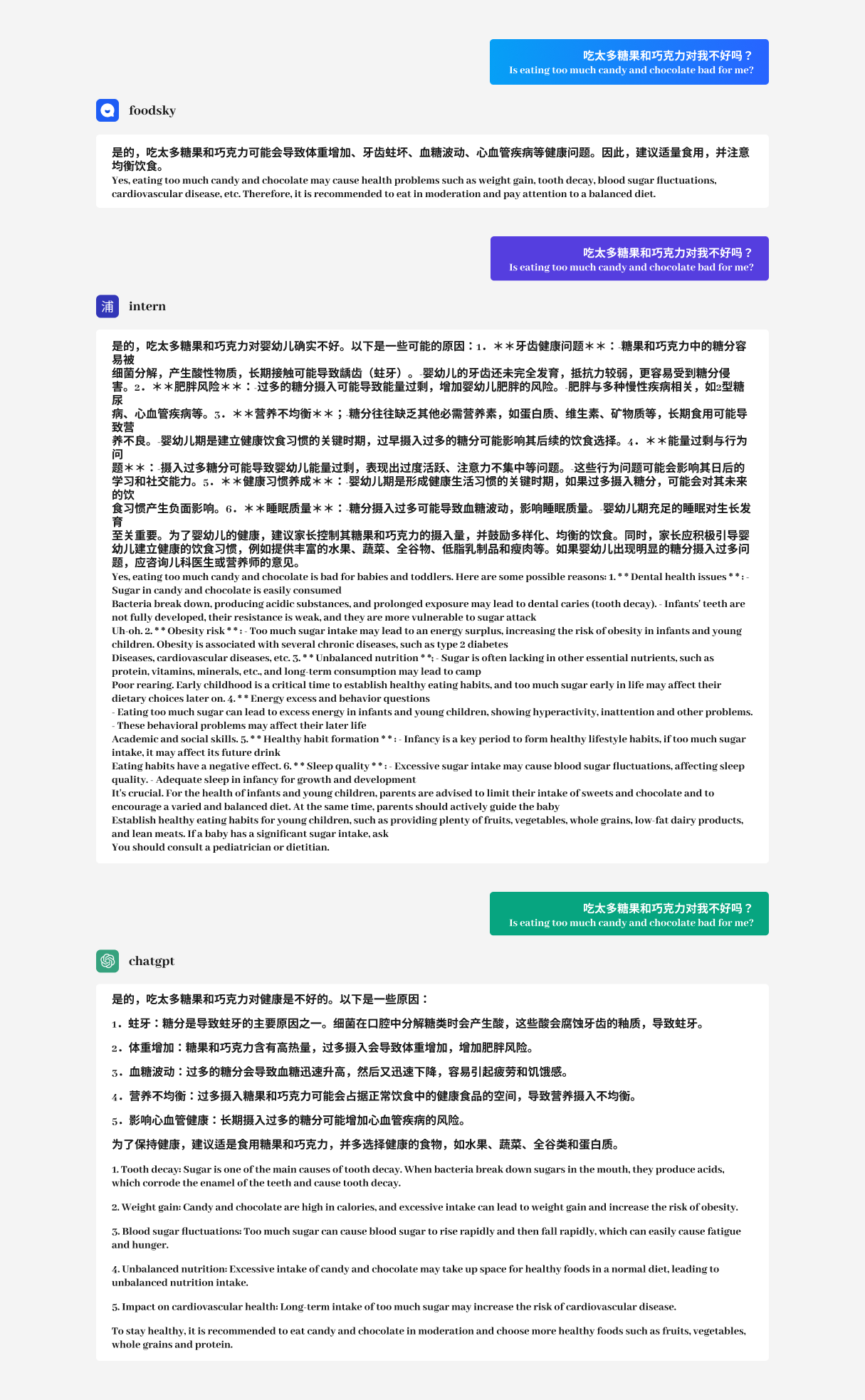}
	\caption{Visualization of dietary education question-answering for the children user group by different models. Our FoodSky gives the most concise and precise answers.}
	\label{fig:kidQA}
\end{figure*}

\begin{figure*}[t]
	\centering
	\includegraphics[width=0.7\textwidth]{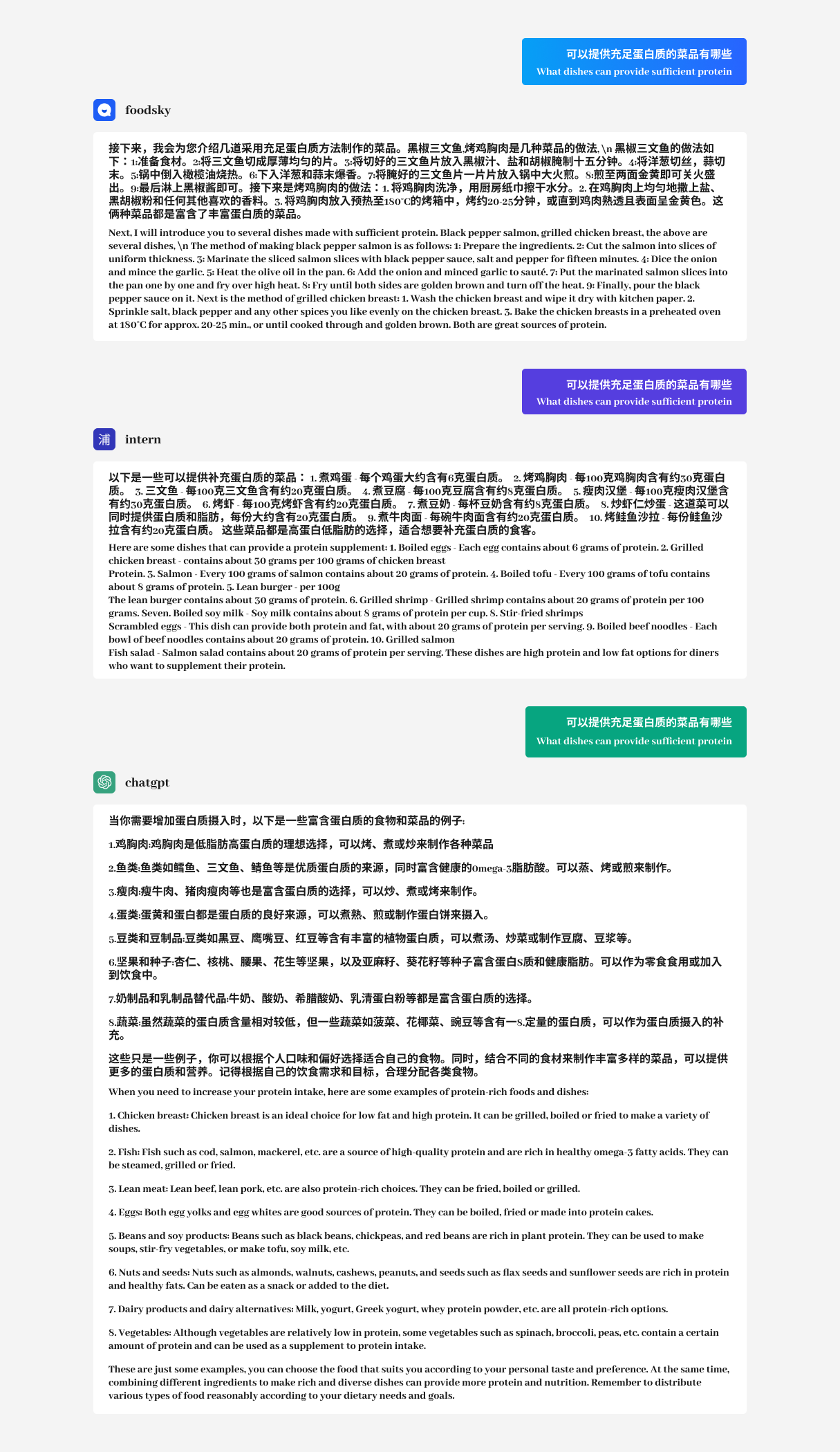}
	\caption{Visualization of nutritional supplement question-answering for the sportspeople user group by different models. Our FoodSky gives the most concise and precise answers.}
	\label{fig:BodyBuilderQA}
\end{figure*}

\begin{figure*}[t]
	\centering
	\includegraphics[width=0.7\textwidth]{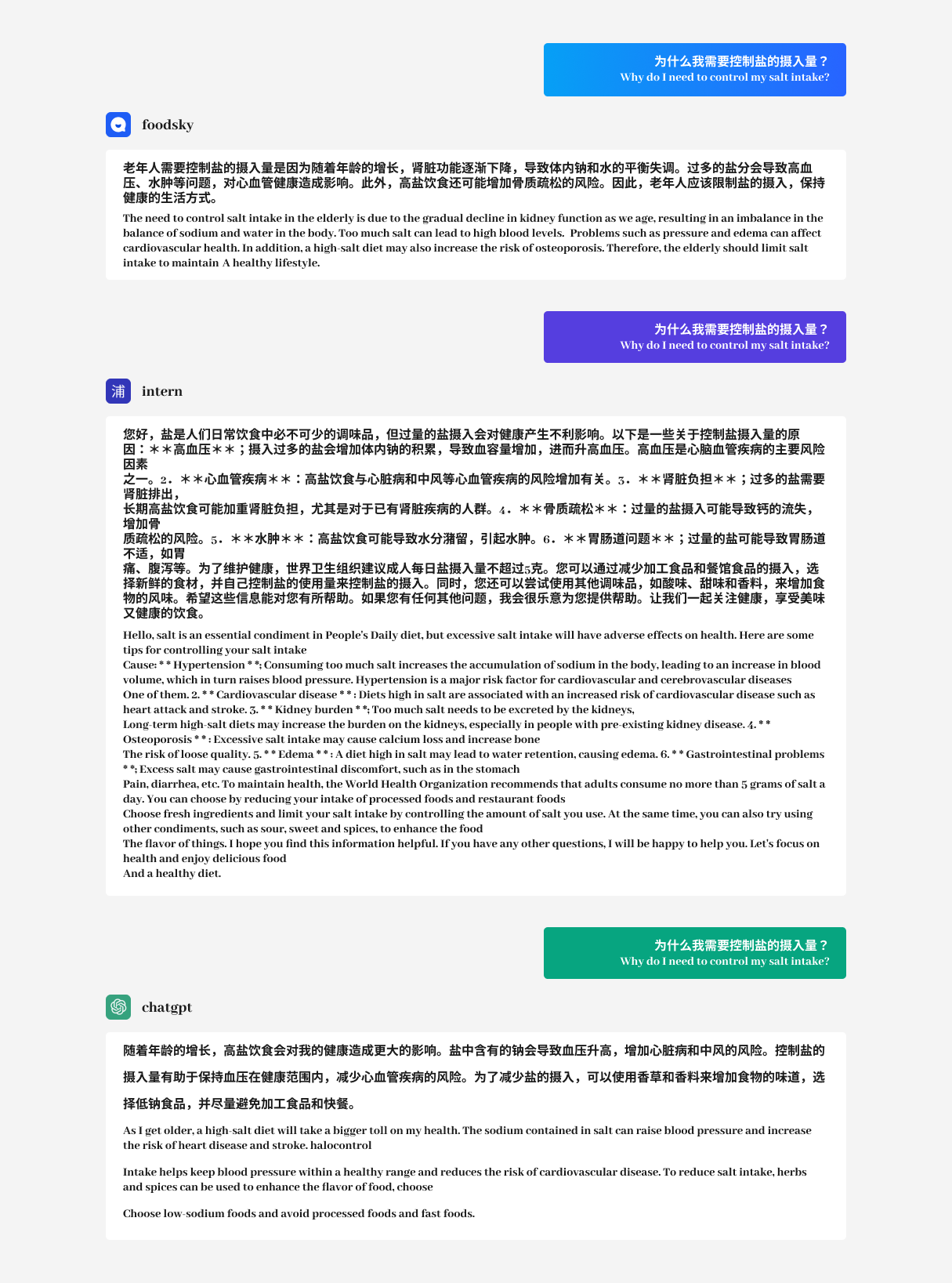}
	\caption{Visualization of health maintenance question-answering for the elderly user group by different models. Our FoodSky gives the most concise and precise answers.}
	\label{fig:SeniorPeopleQA}
\end{figure*}

\begin{figure*}[t]
	\centering
	\includegraphics[width=0.46\textwidth]{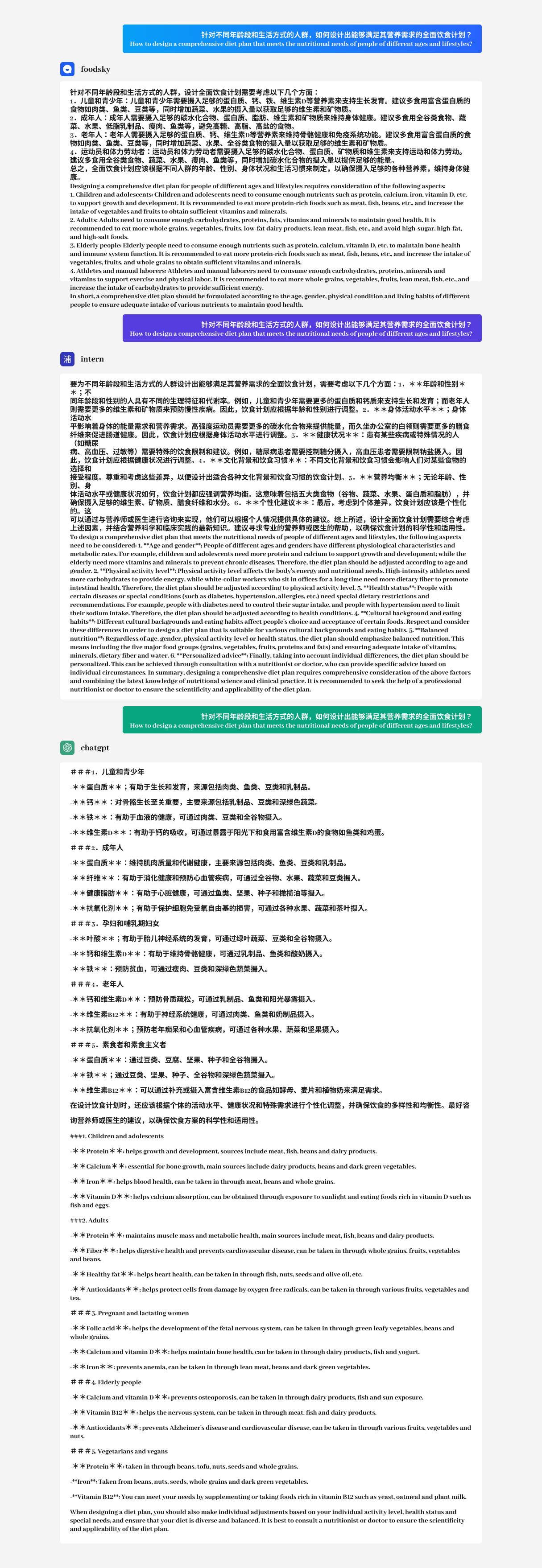}
	\caption{Visualization of academic inspiration question-answering for the healthcare worker user group by different models. Our FoodSky gives the most concise and precise answers.}
	\label{fig:NutritionWorkerQA}
\end{figure*}

\begin{figure*}[t]
	\centering
	\includegraphics[width=0.46\textwidth]{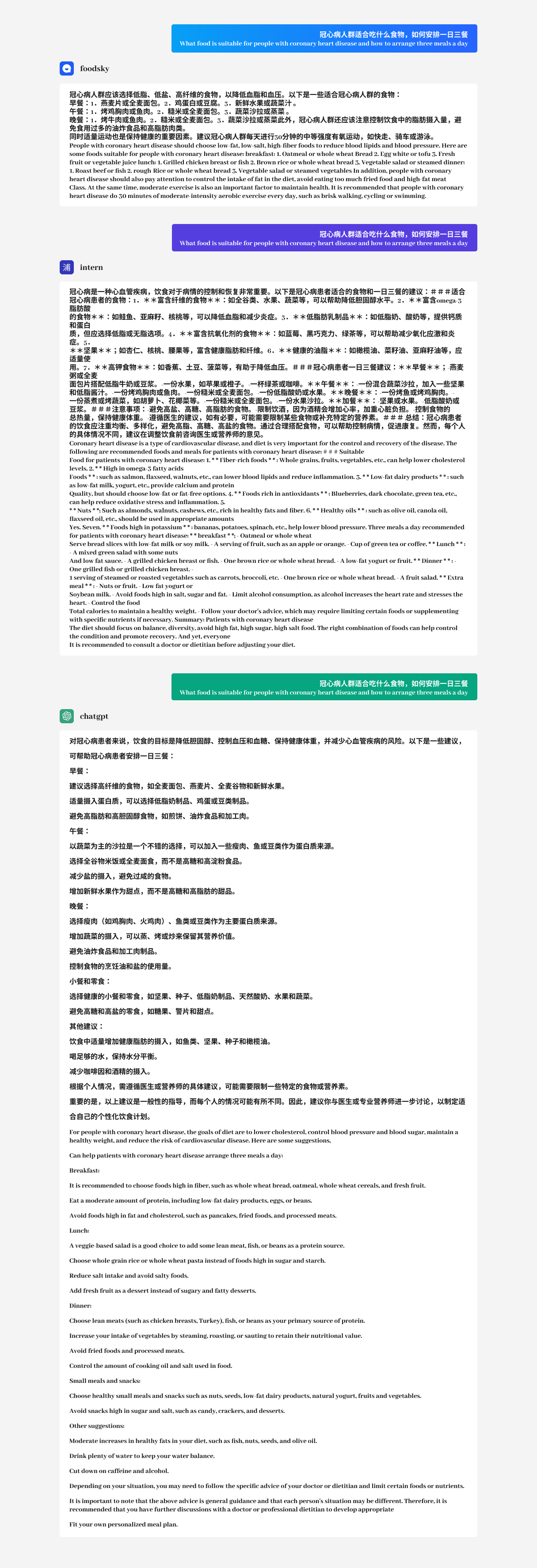}
	\caption{Visualization of health diagnosis question-answering for the chronic patient user group by different models. Our FoodSky gives the most concise and precise answers.}
	\label{fig:diseasePolulationQA}
\end{figure*}

\begin{figure*}[t]
	\centering
	\includegraphics[width=0.46\textwidth]{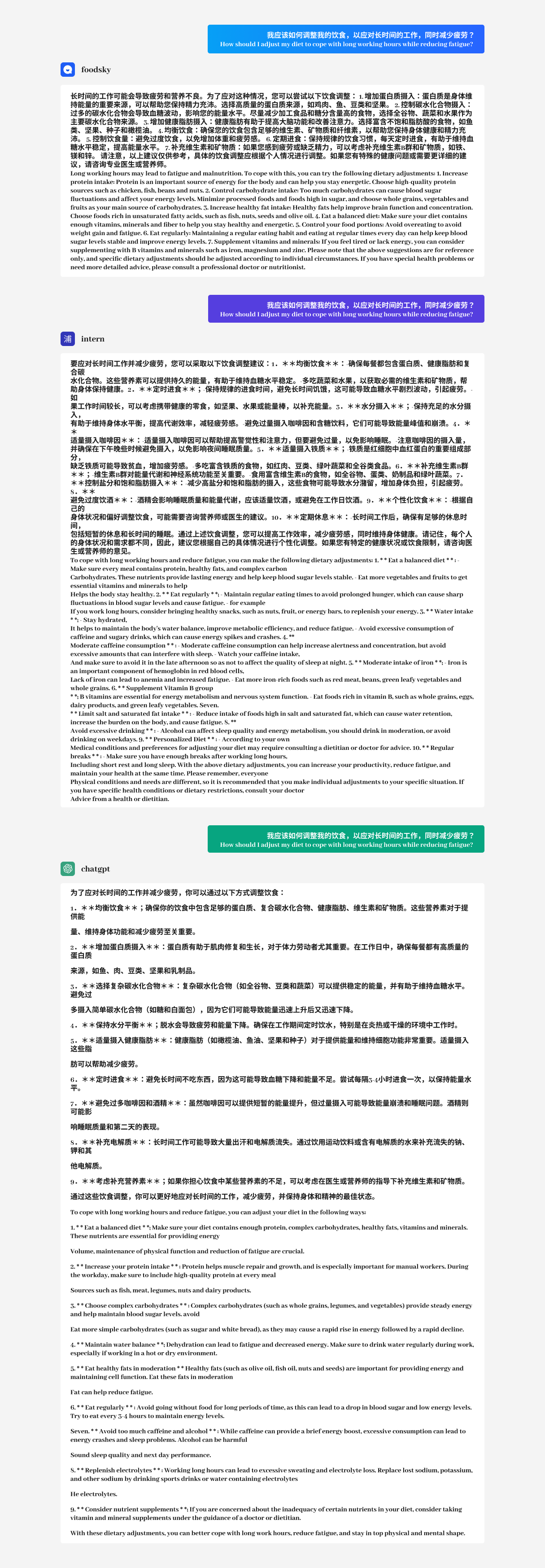}
	\caption{Visualization of meal planning question-answering for the mental worker user group by different models. Our FoodSky gives the most concise and precise answers.}
	\label{fig:BrainWorkerQA}
\end{figure*}

\begin{figure*}[t]
	\centering
	\includegraphics[width=0.6\textwidth]{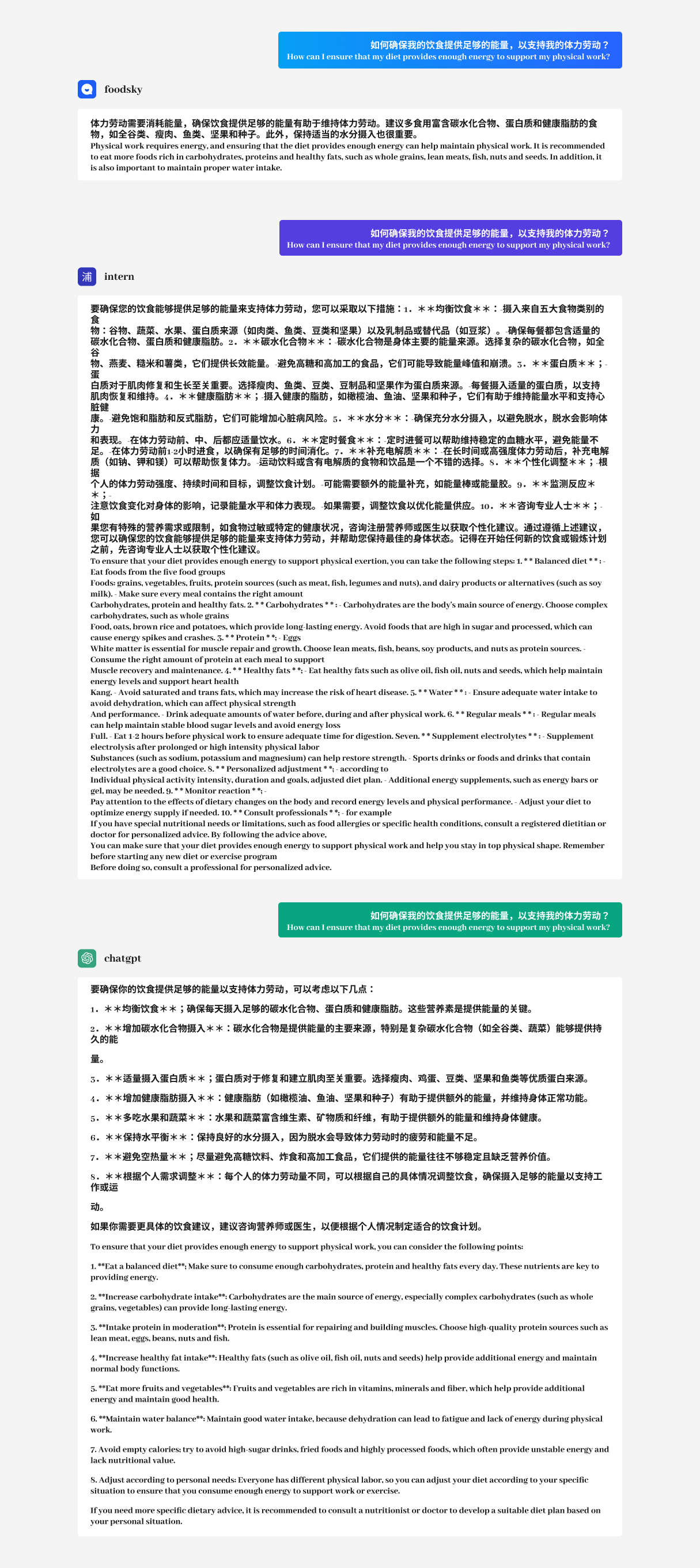}
	\caption{Visualization of nutrition recommendation question-answering for the manual worker user group by different models. Our FoodSky gives the most concise and precise answers.}
	\label{fig:ManualWorkerQA}
\end{figure*}

\begin{figure*}[t]
	\centering
	\includegraphics[width=0.68\textwidth]{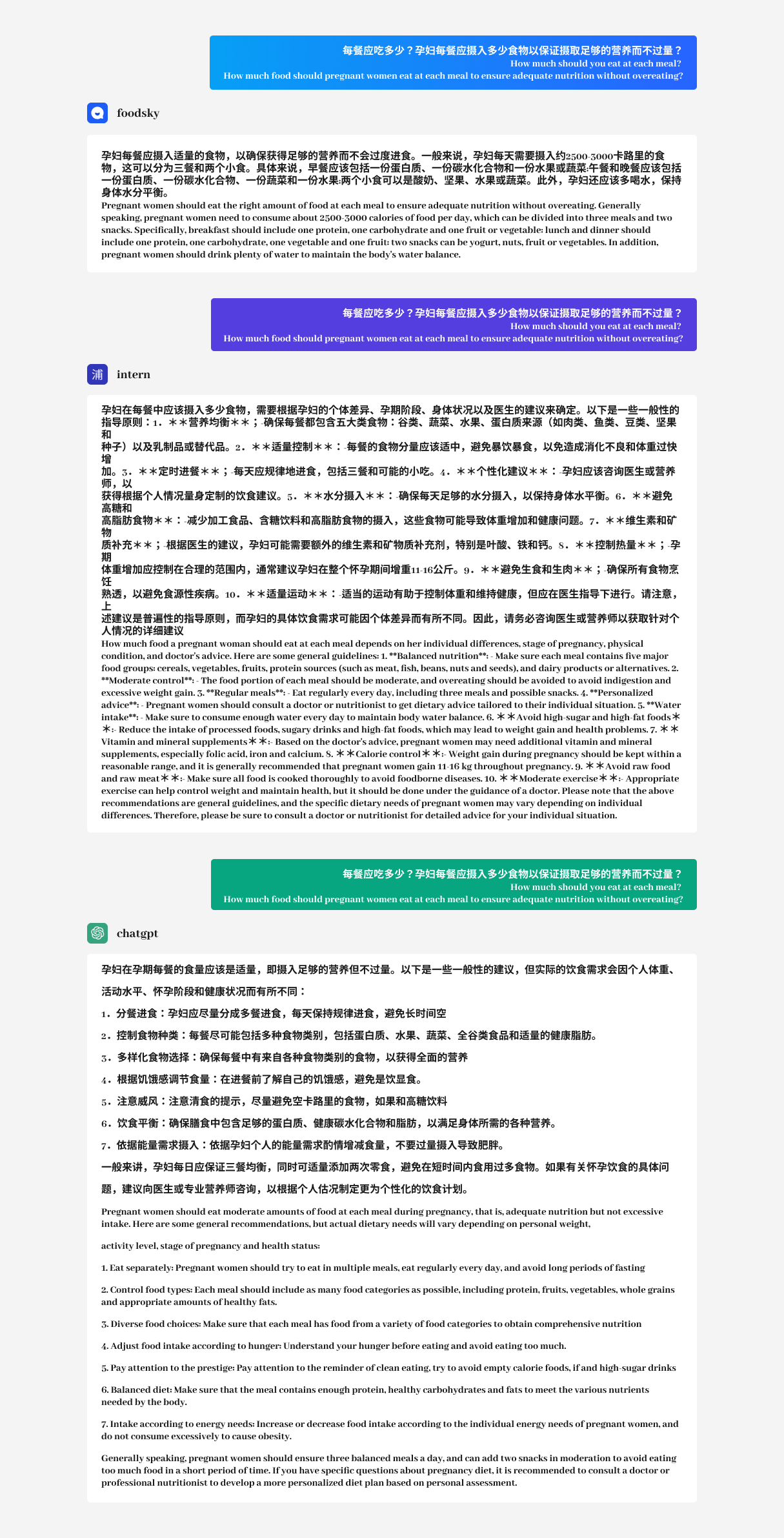}
	\caption{Visualization of dietetic contraindication question-answering for the pregnant woman user group by different models. Our FoodSky gives the most concise and precise answers.}
	\label{fig: PregnantQA}
\end{figure*}


 


\ifCLASSOPTIONcaptionsoff
  \newpage
\fi

\bibliographystyle{IEEEtran}
\bibliography{ref_FoodLMM}

\begin{thebibliography}{100}
\providecommand{\url}[1]{#1}
\csname url@samestyle\endcsname
\providecommand{\newblock}{\relax}
\providecommand{\bibinfo}[2]{#2}
\providecommand{\BIBentrySTDinterwordspacing}{\spaceskip=0pt\relax}
\providecommand{\BIBentryALTinterwordstretchfactor}{4}
\providecommand{\BIBentryALTinterwordspacing}{\spaceskip=\fontdimen2\font plus
\BIBentryALTinterwordstretchfactor\fontdimen3\font minus \fontdimen4\font\relax}
\providecommand{\BIBforeignlanguage}[2]{{%
\expandafter\ifx\csname l@#1\endcsname\relax
\typeout{** WARNING: IEEEtran.bst: No hyphenation pattern has been}%
\typeout{** loaded for the language `#1'. Using the pattern for}%
\typeout{** the default language instead.}%
\else
\language=\csname l@#1\endcsname
\fi
#2}}
\providecommand{\BIBdecl}{\relax}
\BIBdecl

\bibitem{behrens2017evaluating}
P.~Behrens, J.~C. Kiefte-de Jong, T.~Bosker, J.~F. Rodrigues, A.~De~Koning, and A.~Tukker, ``Evaluating the environmental impacts of dietary recommendations,'' \emph{Proceedings of the National Academy of Sciences}, vol. 114, no.~51, pp. 13\,412--13\,417, 2017.

\bibitem{asano2019rising}
Y.~M. Asano and G.~Biermann, ``Rising adoption and retention of meat-free diets in online recipe data,'' \emph{Nature Sustainability}, vol.~2, no.~7, pp. 621--627, 2019.

\bibitem{mehrabi2021global}
Z.~Mehrabi, M.~J. McDowell, V.~Ricciardi, C.~Levers, J.~D. Martinez, N.~Mehrabi, H.~Wittman, N.~Ramankutty, and A.~Jarvis, ``The global divide in data-driven farming,'' \emph{Nature Sustainability}, vol.~4, no.~2, pp. 154--160, 2021.

\bibitem{basso2020digital}
B.~Basso and J.~Antle, ``Digital agriculture to design sustainable agricultural systems,'' \emph{Nature Sustainability}, vol.~3, no.~4, pp. 254--256, 2020.

\bibitem{marin2021recipe1m+}
J.~Mar{\i}n, A.~Biswas, F.~Ofli, N.~Hynes, A.~Salvador, Y.~Aytar, I.~Weber, and A.~Torralba, ``Recipe1m+: A dataset for learning cross-modal embeddings for cooking recipes and food images,'' \emph{IEEE Transactions on Pattern Analysis and Machine Intelligence}, vol.~43, no.~1, pp. 187--203, 2021.

\bibitem{damen2020epic}
D.~Damen, H.~Doughty, G.~M. Farinella \emph{et~al.}, ``The epic-kitchens dataset: Collection, challenges and baselines,'' \emph{IEEE Transactions on Pattern Analysis and Machine Intelligence}, vol.~43, no.~11, pp. 4125--4141, 2020.

\bibitem{siegrist2020consumer}
M.~Siegrist and C.~Hartmann, ``Consumer acceptance of novel food technologies,'' \emph{Nature Food}, vol.~1, no.~6, pp. 343--350, 2020.

\bibitem{popovski2019foodbase}
G.~Popovski, B.~K. Seljak, and T.~Eftimov, ``Foodbase corpus: a new resource of annotated food entities,'' \emph{Database}, vol. 2019, p. baz121, 2019.

\bibitem{thames2021nutrition5k}
Q.~Thames, A.~Karpur, W.~Norris, F.~Xia, L.~Panait, T.~Weyand, and J.~Sim, ``Nutrition5k: Towards automatic nutritional understanding of generic food,'' in \emph{Proceedings of the IEEE/CVF Conference on Computer Vision and Pattern Recognition}, 2021, pp. 8903--8911.

\bibitem{haussmann2019foodkg}
S.~Haussmann, O.~Seneviratne, Y.~Chen, Y.~Ne’eman, J.~Codella, C.-H. Chen, D.~L. McGuinness, and M.~J. Zaki, ``Foodkg: a semantics-driven knowledge graph for food recommendation,'' in \emph{Proceedings of the International Semantic Web Conference}, 2019, pp. 146--162.

\bibitem{min2019survey}
W.~Min, S.~Jiang, L.~Liu, Y.~Rui, and R.~Jain, ``A survey on food computing,'' \emph{ACM Computing Surveys (CSUR)}, vol.~52, no.~5, pp. 1--36, 2019.

\bibitem{grace2021framework}
K.~Grace, S.~Siddiqui, and B.~F. Zaitchik, ``A framework for interdisciplinary research in food systems,'' \emph{Nature Food}, vol.~2, no.~1, pp. 1--3, 2021.

\bibitem{min2023plate}
W.~Min, P.~Zhou, L.~Xu, T.~Liu, T.~Li, M.~Huang, Y.~Jin, Y.~Yi, M.~Wen, S.~Jiang \emph{et~al.}, ``From plate to production: Artificial intelligence in modern consumer-driven food systems,'' \emph{arXiv preprint arXiv:2311.02400}, 2023.

\bibitem{fabregas2019realizing}
R.~Fabregas, M.~Kremer, and F.~Schilbach, ``Realizing the potential of digital development: The case of agricultural advice,'' \emph{Science}, vol. 366, no. 6471, p. eaay3038, 2019.

\bibitem{guo2023wearable}
Y.~Guo, ``Wearable sensors to monitor plant health,'' \emph{Nature Food}, vol.~4, no.~5, pp. 350--350, 2023.

\bibitem{king2017technology}
A.~King, ``Technology: The future of agriculture,'' \emph{Nature}, vol. 544, no. 7651, pp. S21--S23, 2017.

\bibitem{stella2023can}
F.~Stella, C.~Della~Santina, and J.~Hughes, ``How can {LLMs} transform the robotic design process?'' \emph{Nature Machine Intelligence}, pp. 1--4, 2023.

\bibitem{jones2021bubble}
T.~J. Jones, E.~Jambon-Puillet, J.~Marthelot, and P.-T. Brun, ``Bubble casting soft robotics,'' \emph{Nature}, vol. 599, no. 7884, pp. 229--233, 2021.

\bibitem{ummadisingu2022cluttered}
A.~Ummadisingu, K.~Takahashi, and N.~Fukaya, ``Cluttered food grasping with adaptive fingers and synthetic-data trained object detection,'' in \emph{Proceedings of the International Conference on Robotics and Automation (ICRA)}, 2022, pp. 8290--8297.

\bibitem{zoran2019cooking}
A.~Zoran, ``Cooking with computers: The vision of digital gastronomy [point of view],'' \emph{Proceedings of the IEEE}, vol. 107, no.~8, pp. 1467--1473, 2019.

\bibitem{ahnert2013network}
S.~E. Ahnert, ``Network analysis and data mining in food science: the emergence of computational gastronomy,'' \emph{Flavour}, vol.~2, pp. 1--3, 2013.

\bibitem{ahn2011flavor}
Y.-Y. Ahn, S.~E. Ahnert, J.~P. Bagrow, and A.-L. Barab{\'a}si, ``Flavor network and the principles of food pairing,'' \emph{Scientific reports}, vol.~1, no.~1, p. 196, 2011.

\bibitem{althoff2022large}
T.~Althoff, H.~Nilforoshan, J.~Hua, and J.~Leskovec, ``Large-scale diet tracking data reveal disparate associations between food environment and diet,'' \emph{Nature Communications}, vol.~13, no.~1, pp. 1--12, 2022.

\bibitem{afshin2019health}
A.~Afshin, P.~J. Sur, K.~A. Fay, L.~Cornaby, G.~Ferrara, J.~S. Salama, E.~C. Mullany, K.~H. Abate, C.~Abbafati, Z.~Abebe \emph{et~al.}, ``Health effects of dietary risks in 195 countries, 1990--2017: a systematic analysis for the global burden of disease study 2017,'' \emph{The Lancet}, vol. 393, no. 10184, pp. 1958--1972, 2019.

\bibitem{zhou2020cmrdf}
P.~Zhou, C.~Bai, J.~Xia, and S.~Chen, ``Cmrdf: A real-time food alerting system based on multimodal data,'' \emph{IEEE Internet of Things Journal}, vol.~9, no.~9, pp. 6335--6349, 2020.

\bibitem{leng2018network}
R.~I. Leng, ``A network analysis of the propagation of evidence regarding the effectiveness of fat-controlled diets in the secondary prevention of coronary heart disease (chd): Selective citation in reviews,'' \emph{PLoS One}, vol.~13, no.~5, p. e0197716, 2018.

\bibitem{luo2023ingredient}
M.~Luo, W.~Min, Z.~Wang, J.~Song, and S.~Jiang, ``Ingredient prediction via context learning network with class-adaptive asymmetric loss,'' \emph{IEEE Transactions on Image Processing}, 2023.

\bibitem{wang2022ingredient}
Z.~Wang, W.~Min, Z.~Li, L.~Kang, X.~Wei, X.~Wei, and S.~Jiang, ``Ingredient-guided region discovery and relationship modeling for food category-ingredient prediction,'' \emph{IEEE Transactions on Image Processing}, vol.~31, pp. 5214--5226, 2022.

\bibitem{wen2023multi}
M.~Wen, J.~Song, W.~Min, W.~Xiao, L.~Han, and S.~Jiang, ``Multi-state ingredient recognition via adaptive multi-centric network,'' \emph{IEEE Transactions on Industrial Informatics}, 2023.

\bibitem{zhu2019r2gan}
B.~Zhu, C.-W. Ngo, J.~Chen, and Y.~Hao, ``R2gan: Cross-modal recipe retrieval with generative adversarial network,'' in \emph{Proceedings of the IEEE/CVF Conference on Computer Vision and Pattern Recognition}, 2019, pp. 11\,477--11\,486.

\bibitem{salvador2019inverse}
A.~Salvador, M.~Drozdzal, X.~Gir{\'o}-i Nieto, and A.~Romero, ``Inverse cooking: Recipe generation from food images,'' in \emph{Proceedings of the IEEE/CVF Conference on Computer Vision and Pattern Recognition}, 2019, pp. 10\,453--10\,462.

\bibitem{wang2022review}
W.~Wang, W.~Min, T.~Li, X.~Dong, H.~Li, and S.~Jiang, ``A review on vision-based analysis for automatic dietary assessment,'' \emph{Trends in Food Science \& Technology}, vol. 122, pp. 223--237, 2022.

\bibitem{li2023deep}
T.~Li, W.~Wei, S.~Xing, W.~Min, C.~Zhang, and S.~Jiang, ``Deep learning-based near-infrared hyperspectral imaging for food nutrition estimation,'' \emph{Foods}, vol.~12, no.~17, p. 3145, 2023.

\bibitem{shao2023vision}
W.~Shao, W.~Min, S.~Hou, M.~Luo, T.~Li, Y.~Zheng, and S.~Jiang, ``Vision-based food nutrition estimation via rgb-d fusion network,'' \emph{Food Chemistry}, vol. 424, p. 136309, 2023.

\bibitem{liu2024convolution}
Y.~Liu, W.~Min, S.~Jiang, and Y.~Rui, ``Convolution-enhanced bi-branch adaptive transformer with cross-task interaction for food category and ingredient recognition,'' \emph{IEEE Transactions on Image Processing}, 2024.

\bibitem{gaupp2021food}
F.~Gaupp, C.~Ruggeri~Laderchi, H.~Lotze-Campen, F.~DeClerck, B.~L. Bodirsky, S.~Lowder, A.~Popp, R.~Kanbur, O.~Edenhofer, R.~Nugent \emph{et~al.}, ``Food system development pathways for healthy, nature-positive and inclusive food systems,'' \emph{Nature Food}, vol.~2, no.~12, pp. 928--934, 2021.

\bibitem{min2023large}
W.~Min, Z.~Wang, Y.~Liu, M.~Luo, L.~Kang, X.~Wei, X.~Wei, and S.~Jiang, ``Large scale visual food recognition,'' \emph{IEEE Transactions on Pattern Analysis and Machine Intelligence}, vol.~45, no.~8, pp. 9932--9949, 2023.

\bibitem{hadi2023survey}
M.~U. Hadi, R.~Qureshi, A.~Shah \emph{et~al.}, ``A survey on large language models: Applications, challenges, limitations, and practical usage,'' \emph{Authorea Preprints}, 2023.

\bibitem{zhou2023survey}
H.~Zhou, B.~Gu, X.~Zou \emph{et~al.}, ``A survey of large language models in medicine: Progress, application, and challenge,'' \emph{arXiv preprint arXiv:2311.05112}, 2023.

\bibitem{yang2024zhongjing}
S.~Yang, H.~Zhao, S.~Zhu, G.~Zhou, H.~Xu, Y.~Jia, and H.~Zan, ``Zhongjing: Enhancing the chinese medical capabilities of large language model through expert feedback and real-world multi-turn dialogue,'' in \emph{Proceedings of the AAAI Conference on Artificial Intelligence}, 2024, pp. 19\,368--19\,376.

\bibitem{wang2024large}
S.~Wang, T.~Xu, H.~Li, C.~Zhang, J.~Liang, J.~Tang, P.~S. Yu, and Q.~Wen, ``Large language models for education: A survey and outlook,'' \emph{arXiv preprint arXiv:2403.18105}, 2024.

\bibitem{wardat2023chatgpt}
Y.~Wardat, M.~A. Tashtoush, R.~AlAli, and A.~M. Jarrah, ``Chatgpt: A revolutionary tool for teaching and learning mathematics,'' \emph{Eurasia Journal of Mathematics, Science and Technology Education}, vol.~19, no.~7, p. em2286, 2023.

\bibitem{lee2024survey}
J.~Lee, N.~Stevens, S.~C. Han, and M.~Song, ``A survey of large language models in finance (finllms),'' \emph{arXiv preprint arXiv:2402.02315}, 2024.

\bibitem{wu2023bloomberggpt}
S.~Wu, O.~Irsoy, S.~Lu, V.~Dabravolski, M.~Dredze, S.~Gehrmann, P.~Kambadur, D.~Rosenberg, and G.~Mann, ``Bloomberggpt: A large language model for finance,'' \emph{arXiv preprint arXiv:2303.17564}, 2023.

\bibitem{xu2023baize}
C.~Xu, D.~Guo, N.~Duan, and J.~McAuley, ``Baize: An open-source chat model with parameter-efficient tuning on self-chat data,'' in \emph{Proceedings of the 2023 Conference on Empirical Methods in Natural Language Processing}, 2023.

\bibitem{zhang2023huatuogpt}
H.~Zhang, J.~Chen, F.~Jiang \emph{et~al.}, ``Huatuogpt, towards taming language model to be a doctor,'' in \emph{Proceedings of the 2023 Conference on Empirical Methods in Natural Language Processing}, 2023.

\bibitem{qi2023foodgpt}
Z.~Qi, Y.~Yu, M.~Tu, J.~Tan, and Y.~Huang, ``Foodgpt: A large language model in food testing domain with incremental pre-training and knowledge graph prompt,'' \emph{arXiv preprint arXiv:2308.10173}, 2023.

\bibitem{cunningham2023foodgpt}
R.~Cunningham, S.~Farnum, S.~Kang, and R.~Saxena, ``Foodgpt: Amachine learning approach to ingredient substitution and recipe recommendation,'' 2023.

\bibitem{yin2023foodlmm}
Y.~Yin, H.~Qi, B.~Zhu, J.~Chen, Y.-G. Jiang, and C.-W. Ngo, ``Foodlmm: A versatile food assistant using large multi-modal model,'' \emph{arXiv preprint arXiv:2312.14991}, 2023.

\bibitem{zhou2024does}
L.~Zhou, T.~Karidi, N.~Garneau, Y.~Cao, W.~Liu, W.~Chen, and D.~Hershcovich, ``Does mapo tofu contain coffee? probing llms for food-related cultural knowledge,'' \emph{arXiv preprint arXiv:2404.06833}, 2024.

\bibitem{internlm2}
H.~C. Zheng~Cai, Maosong~Cao \emph{et~al.}, ``Internlm2 technical report,'' 2024.

\bibitem{chatgpt}
L.~Ouyang, J.~Wu, X.~Jiang \emph{et~al.}, ``Training language models to follow instructions with human feedback,'' 2022.

\bibitem{pinker1995language}
S.~Pinker, \emph{The language instinct: The new science of language and mind}, 1995, vol. 7529.

\bibitem{turing2009computing}
A.~M. Turing, \emph{Computing machinery and intelligence}, 2009.

\bibitem{zhao2023survey}
W.~X. Zhao, K.~Zhou, J.~Li \emph{et~al.}, ``A survey of large language models,'' \emph{arXiv preprint arXiv:2303.18223}, 2023.

\bibitem{shanahan2024talking}
M.~Shanahan, ``Talking about large language models,'' \emph{Communications of the ACM}, vol.~67, no.~2, pp. 68--79, 2024.

\bibitem{zhou2023chatgpt}
J.~Zhou, P.~Ke, X.~Qiu, M.~Huang, and J.~Zhang, ``Chatgpt: potential, prospects, and limitations,'' \emph{Frontiers of Information Technology \& Electronic Engineering}, pp. 1--6, 2023.

\bibitem{chowdhery2023palm}
A.~Chowdhery, S.~Narang, J.~Devlin \emph{et~al.}, ``Palm: Scaling language modeling with pathways,'' \emph{Journal of Machine Learning Research}, vol.~24, no. 240, pp. 1--113, 2023.

\bibitem{taylor2022galactica}
R.~Taylor, M.~Kardas, G.~Cucurull, T.~Scialom, A.~Hartshorn, E.~Saravia, A.~Poulton, V.~Kerkez, and R.~Stojnic, ``Galactica: A large language model for science,'' \emph{arXiv preprint arXiv:2211.09085}, 2022.

\bibitem{touvron2023llama}
H.~Touvron, T.~Lavril, G.~Izacard \emph{et~al.}, ``Llama: Open and efficient foundation language models,'' \emph{arXiv preprint arXiv:2302.13971}, 2023.

\bibitem{jakubik2024data}
J.~Jakubik, M.~V{\"o}ssing, N.~K{\"u}hl, J.~Walk, and G.~Satzger, ``Data-centric artificial intelligence,'' \emph{Business \& Information Systems Engineering}, pp. 1--9, 2024.

\bibitem{openai-blog}
S.~Altman, ``{Planning for AGI and Beyond},'' \url{https://openai.com/blog/planning-for-agi-and-beyond}, 2023.

\bibitem{bubeck2023sparks}
S.~Bubeck, V.~Chandrasekaran, R.~Eldan \emph{et~al.}, ``Sparks of artificial general intelligence: Early experiments with gpt-4,'' \emph{arXiv preprint arXiv:2303.12712}, 2023.

\bibitem{huang2024language}
S.~Huang, L.~Dong, W.~Wang \emph{et~al.}, ``Language is not all you need: Aligning perception with language models,'' in \emph{Proceedings of the Advances in Neural Information Processing Systems}, 2024.

\bibitem{cao2023comprehensive}
Y.~Cao, S.~Li, Y.~Liu, Z.~Yan, Y.~Dai, P.~S. Yu, and L.~Sun, ``A comprehensive survey of ai-generated content (aigc): A history of generative ai from gan to chatgpt,'' \emph{arXiv preprint arXiv:2303.04226}, 2023.

\bibitem{driess2023palm}
D.~Driess, F.~Xia, M.~S. Sajjadi \emph{et~al.}, ``Palm-e: An embodied multimodal language model,'' \emph{arXiv preprint arXiv:2303.03378}, 2023.

\bibitem{wu2023visual}
C.~Wu, S.~Yin, W.~Qi, X.~Wang, Z.~Tang, and N.~Duan, ``Visual chatgpt: Talking, drawing and editing with visual foundation models,'' \emph{arXiv preprint arXiv:2303.04671}, 2023.

\bibitem{gilson2023does}
A.~Gilson, C.~W. Safranek, T.~Huang \emph{et~al.}, ``How does chatgpt perform on the united states medical licensing examination (usmle)? the implications of large language models for medical education and knowledge assessment,'' \emph{JMIR Medical Education}, vol.~9, no.~1, p. e45312, 2023.

\bibitem{kung2023performance}
T.~H. Kung, M.~Cheatham, A.~Medenilla \emph{et~al.}, ``Performance of chatgpt on usmle: potential for ai-assisted medical education using large language models,'' \emph{PLoS digital health}, vol.~2, no.~2, p. e0000198, 2023.

\bibitem{duong2024analysis}
D.~Duong and B.~D. Solomon, ``Analysis of large-language model versus human performance for genetics questions,'' \emph{European Journal of Human Genetics}, vol.~32, no.~4, pp. 466--468, 2024.

\bibitem{gan2023large}
W.~Gan, Z.~Qi, J.~Wu, and J.~C.-W. Lin, ``Large language models in education: Vision and opportunities,'' in \emph{Proceedings of the 2023 IEEE International Conference on Big Data}, 2023, pp. 4776--4785.

\bibitem{fraiwan2023review}
M.~Fraiwan and N.~Khasawneh, ``A review of chatgpt applications in education, marketing, software engineering, and healthcare: Benefits, drawbacks, and research directions,'' \emph{arXiv preprint arXiv:2305.00237}, 2023.

\bibitem{xie2023pixiu}
Q.~Xie, W.~Han, X.~Zhang, Y.~Lai, M.~Peng, A.~Lopez-Lira, and J.~Huang, ``Pixiu: A large language model, instruction data and evaluation benchmark for finance,'' \emph{arXiv preprint arXiv:2306.05443}, 2023.

\bibitem{yang2023investlm}
Y.~Yang, Y.~Tang, and K.~Y. Tam, ``Investlm: A large language model for investment using financial domain instruction tuning,'' \emph{arXiv preprint arXiv:2309.13064}, 2023.

\bibitem{yang2023fingpt}
H.~Yang, X.-Y. Liu, and C.~D. Wang, ``Fingpt: Open-source financial large language models,'' \emph{arXiv preprint arXiv:2306.06031}, 2023.

\bibitem{zhou2024synthesizing}
P.~Zhou, W.~Min, J.~Song, Y.~Zhang, and S.~Jiang, ``Synthesizing knowledge-enhanced features for real-world zero-shot food detection,'' \emph{IEEE Transactions on Image Processing}, 2024.

\bibitem{min2022applications}
W.~Min, C.~Liu, L.~Xu, and S.~Jiang, ``Applications of knowledge graphs for food science and industry,'' \emph{Patterns}, vol.~3, no.~5, 2022.

\bibitem{menichetti2023machine}
G.~Menichetti, B.~Ravandi, D.~Mozaffarian, and A.-L. Barab{\'a}si, ``Machine learning prediction of the degree of food processing,'' \emph{Nature Communications}, vol.~14, no.~1, p. 2312, 2023.

\bibitem{yang2021integrated}
R.~Yang, Z.~Wang, and J.~Chen, ``An integrated approach of mechanistic-modeling and machine-learning for thickness optimization of frozen microwaveable foods,'' \emph{Foods}, vol.~10, no.~4, p. 763, 2021.

\bibitem{min2019food}
W.~Min, S.~Jiang, and R.~Jain, ``Food recommendation: Framework, existing solutions, and challenges,'' \emph{IEEE Transactions on Multimedia}, vol.~22, no.~10, pp. 2659--2671, 2019.

\bibitem{chen2021personalized}
Y.~Chen, A.~Subburathinam, C.-H. Chen, and M.~J. Zaki, ``Personalized food recommendation as constrained question answering over a large-scale food knowledge graph,'' in \emph{Proceedings of the ACM International Conference on Web Search and Data Mining}, 2021, pp. 544--552.

\bibitem{singh2022conversational}
E.~Singh, A.~Bompelli, R.~Wan, J.~Bian, S.~Pakhomov, and R.~Zhang, ``A conversational agent system for dietary supplements use,'' \emph{BMC medical informatics and decision making}, vol.~22, p. 153, 2022.

\bibitem{lin2014content}
C.-J. Lin, T.-T. Kuo, and S.-D. Lin, ``A content-based matrix factorization model for recipe recommendation,'' in \emph{Proceedings of the Advances in Knowledge Discovery and Data Mining}, 2014, pp. 560--571.

\bibitem{zhang2016exploiting}
F.~Zhang, N.~J. Yuan, K.~Zheng, D.~Lian, X.~Xie, and Y.~Rui, ``Exploiting dining preference for restaurant recommendation,'' in \emph{Proceedings of the International Conference on World Wide Web}, 2016, pp. 725--735.

\bibitem{cui2023construction}
J.~Cui, X.~Zhang, and D.~Zheng, ``Construction of recipe knowledge graph based on user knowledge demands,'' \emph{Journal of Information Science}, 2023.

\bibitem{li2023health}
D.~Li, M.~J. Zaki, and C.-h. Chen, ``Health-guided recipe recommendation over knowledge graphs,'' \emph{Journal of Web Semantics}, vol.~75, p. 100743, 2023.

\bibitem{chu2017hybrid}
W.-T. Chu and Y.-L. Tsai, ``A hybrid recommendation system considering visual information for predicting favorite restaurants,'' \emph{World Wide Web}, vol.~20, pp. 1313--1331, 2017.

\bibitem{asani2021restaurant}
E.~Asani, H.~Vahdat-Nejad, and J.~Sadri, ``Restaurant recommender system based on sentiment analysis,'' \emph{Machine Learning with Applications}, vol.~6, p. 100114, 2021.

\bibitem{ling2022following}
Y.~Ling, J.-Y. Nie, D.~Nielsen, B.~Kn{\"a}uper, N.~Yang, and L.~Dub{\'e}, ``Following good examples-health goal-oriented food recommendation based on behavior data,'' in \emph{Proceedings of the ACM Web Conference}, 2022, pp. 3745--3754.

\bibitem{ribeiro2018souschef}
D.~Ribeiro, J.~Ribeiro, M.~J.~M. Vasconcelos, E.~F. Vieira, and A.~C. de~Barros, ``Souschef: improved meal recommender system for portuguese older adults,'' in \emph{Proceedings of the Information and Communication Technologies for Ageing Well and e-Health}, 2018, pp. 107--126.

\bibitem{salvador2017learning}
A.~Salvador, N.~Hynes, Y.~Aytar, J.~Marin, F.~Ofli, I.~Weber, and A.~Torralba, ``Learning cross-modal embeddings for cooking recipes and food images,'' in \emph{Proceedings of the IEEE Conference on Computer Vision and Pattern Recognition}, 2017, pp. 3020--3028.

\bibitem{salvador2021revamping}
A.~Salvador, E.~Gundogdu, L.~Bazzani, and M.~Donoser, ``Revamping cross-modal recipe retrieval with hierarchical transformers and self-supervised learning,'' in \emph{Proceedings of the IEEE/CVF Conference on Computer Vision and Pattern Recognition}, 2021, pp. 15\,475--15\,484.

\bibitem{wang2021cross}
H.~Wang, D.~Sahoo, C.~Liu, K.~Shu, P.~Achananuparp, E.-p. Lim, and S.~C. Hoi, ``Cross-modal food retrieval: learning a joint embedding of food images and recipes with semantic consistency and attention mechanism,'' \emph{IEEE Transactions on Multimedia}, vol.~24, pp. 2515--2525, 2021.

\bibitem{sugiyama2021cross}
Y.~Sugiyama and K.~Yanai, ``Cross-modal recipe embeddings by disentangling recipe contents and dish styles,'' in \emph{Proceedings of the ACM International Conference on Multimedia}, 2021, pp. 2501--2509.

\bibitem{guerrero2020cross}
R.~Guerrero, H.~X. Pham, and V.~Pavlovic, ``Cross-modal retrieval and synthesis (x-mrs): closing the modality gap in shared representation learning,'' \emph{arXiv preprint arXiv:2012.01345}, 2020.

\bibitem{qarajeh2023ai}
A.~Qarajeh, S.~Tangpanithandee, C.~Thongprayoon \emph{et~al.}, ``Ai-powered renal diet support: Performance of chatgpt, bard ai, and bing chat,'' \emph{Clinics and Practice}, vol.~13, no.~5, pp. 1160--1172, 2023.

\bibitem{liu2024visual}
H.~Liu, C.~Li, Q.~Wu, and Y.~J. Lee, ``Visual instruction tuning,'' 2024.

\bibitem{nag2023integrative}
N.~Nag, H.~Oh, M.~Tang, M.~Shi, and R.~Jain, ``Integrative multi-modal computing for personal health navigation,'' in \emph{Proceedings of the 2023 ACM International Conference on Multimedia Retrieval}, 2023, pp. 1--9.

\bibitem{li2023towards}
Z.~Li, X.~Zhang, Y.~Zhang, D.~Long, P.~Xie, and M.~Zhang, ``Towards general text embeddings with multi-stage contrastive learning,'' \emph{arXiv preprint arXiv:2308.03281}, 2023.

\bibitem{bge_embedding}
S.~Xiao, Z.~Liu, P.~Zhang, and N.~Muennighoff, ``C-pack: Packaged resources to advance general chinese embedding,'' 2023.

\bibitem{intsigacge}
\BIBentryALTinterwordspacing
INTSIG, ``acge\_text\_embedding,'' 2024. [Online]. Available: \url{https://huggingface.co/aspire/acge_text_embedding}
\BIBentrySTDinterwordspacing

\bibitem{yang2023baichuan}
A.~Yang, B.~Xiao, B.~Wang \emph{et~al.}, ``Baichuan 2: Open large-scale language models,'' \emph{arXiv preprint arXiv:2309.10305}, 2023.

\bibitem{infgradstella}
\BIBentryALTinterwordspacing
INFGRAD, ``stella-mrl-large-zh-v3.5,'' 2024. [Online]. Available: \url{https://huggingface.co/infgrad/stella-mrl-large-zh-v3.5-1792d}
\BIBentrySTDinterwordspacing

\bibitem{infgradpuff}
\BIBentryALTinterwordspacing
------, ``puff-base-v1,'' 2024. [Online]. Available: \url{https://huggingface.co/infgrad/puff-base-v1}
\BIBentrySTDinterwordspacing

\bibitem{maas2013rectifier}
A.~L. Maas, A.~Y. Hannun, A.~Y. Ng \emph{et~al.}, ``Rectifier nonlinearities improve neural network acoustic models,'' in \emph{Proceedings of the International Conference on Machine Learning}, 2013, pp. 3--8.

\bibitem{10.1162/tacl_a_00530}
S.~Siriwardhana, R.~Weerasekera, E.~Wen, T.~Kaluarachchi, R.~Rana, and S.~Nanayakkara, ``{Improving the Domain Adaptation of Retrieval Augmented Generation (RAG) Models for Open Domain Question Answering},'' \emph{Transactions of the Association for Computational Linguistics}, vol.~11, pp. 1--17, 2023.

\bibitem{chatglm1}
A.~Zeng, X.~Liu, Z.~Du \emph{et~al.}, ``Glm-130b: An open bilingual pre-trained model,'' \emph{arXiv preprint arXiv:2210.02414}, 2022.

\bibitem{mistral}
A.~Q. Jiang, A.~Sablayrolles, A.~Mensch \emph{et~al.}, ``Mistral 7b,'' 2023.

\bibitem{vicuna2}
\BIBentryALTinterwordspacing
W.-L. Chiang, Z.~Li, Z.~Lin \emph{et~al.}, ``Vicuna: An open-source chatbot impressing gpt-4 with 90\%* chatgpt quality,'' 2023. [Online]. Available: \url{https://lmsys.org/blog/2023-03-30-vicuna/}
\BIBentrySTDinterwordspacing

\bibitem{cllama}
Y.~Cui, Z.~Yang, and X.~Yao, ``Efficient and effective text encoding for chinese llama and alpaca,'' \emph{arXiv preprint arXiv:2304.08177}, 2023.

\bibitem{qwen}
Y.~C. Jinze~Bai, Shuai~Bai \emph{et~al.}, ``Qwen technical report,'' \emph{arXiv preprint arXiv:2309.16609}, 2023.

\bibitem{hu2021lora}
E.~J. Hu, Y.~Shen, P.~Wallis, Z.~Allen-Zhu, Y.~Li, S.~Wang, L.~Wang, and W.~Chen, ``Lora: Low-rank adaptation of large language models,'' 2021.

\bibitem{adam2014}
D.~Kingma and J.~Ba, ``Adam: A method for stochastic optimization,'' in \emph{Proceedings of the International Conference on Learning Representations}, 2015, pp. 1--15.

\bibitem{chatglm2}
Z.~Du, Y.~Qian, X.~Liu, M.~Ding, J.~Qiu, Z.~Yang, and J.~Tang, ``Glm: General language model pretraining with autoregressive blank infilling,'' in \emph{Proceedings of the Annual Meeting of the Association for Computational Linguistics (Volume 1: Long Papers)}, 2022, pp. 320--335.

\bibitem{vicuna1}
L.~Zheng, W.-L. Chiang, Y.~Sheng \emph{et~al.}, ``Judging llm-as-a-judge with mt-bench and chatbot arena,'' 2023.

\bibitem{papineni2002bleu}
K.~Papineni, S.~Roukos, T.~Ward, and W.-J. Zhu, ``{B}leu: a method for automatic evaluation of machine translation,'' in \emph{Proceedings of the Annual Meeting of the Association for Computational Linguistics}.\hskip 1em plus 0.5em minus 0.4em\relax Association for Computational Linguistics, 2002, pp. 311--318.

\bibitem{lin2004rouge}
C.-Y. Lin, ``{ROUGE}: A package for automatic evaluation of summaries,'' in \emph{Proceedings of the Workshop on Text Summarization Branches Out}, 2004, pp. 74--81.

\bibitem{wu2016googles}
Y.~Wu, M.~Schuster \emph{et~al.}, ``Google's neural machine translation system: Bridging the gap between human and machine translation,'' 2016.

\bibitem{li2016diversity}
J.~Li, M.~Galley, C.~Brockett, J.~Gao, and B.~Dolan, ``A diversity-promoting objective function for neural conversation models,'' in \emph{Proceedings of the 2016 Conference of the North {A}merican Chapter of the Association for Computational Linguistics: Human Language Technologies}, 2016, pp. 110--119.

\end{thebibliography}

\vfill

\end{document}